\documentclass[10pt,twocolumn,letterpaper]{article}

\usepackage{cvpr}
\usepackage{times}
\usepackage{epsfig}
\usepackage{graphicx}
\usepackage{amsmath}
\usepackage{amssymb}

\usepackage[font=small,labelsep=period]{caption}
\usepackage{subcaption}
\usepackage{multirow,booktabs}
\usepackage{adjustbox}
\usepackage[usenames,dvipsnames,svgnames,table]{xcolor}
\usepackage{soul}
\usepackage[acronym]{glossaries}
\newacronym{sfm}{SfM}{Structure from Motion}
\newacronym{mst}{MST}{Maximum Spanning Tree}
\newacronym{irls}{IRLS}{Iteratively Reweighted Least Squares}

\usepackage[breaklinks=true,bookmarks=false]{hyperref}

\cvprfinalcopy 
\def\cvprPaperID{****} 

\ifcvprfinal\fi
\begin{document}

\title{Progressive Structure from Motion}

\author{
Alex Locher\textsuperscript{1}
\and Michal Havlena\textsuperscript{1}
\and Luc Van Gool\textsuperscript{1,2}
\and \textsuperscript{1} Computer Vision Laboratory, ETH Zurich, Switzerland
\and \textsuperscript{2} VISICS, KU Leuven, Belgium
}

\maketitle

\begin{abstract}
Structure from Motion or the sparse 3D reconstruction out of individual photos is a long studied topic in computer vision.
Yet none of the existing reconstruction pipelines fully addresses a progressive scenario where images are only getting available during the reconstruction process and intermediate results are delivered to the user.
Incremental pipelines are capable of growing a 3D model but often get stuck in local minima due to wrong (binding) decisions taken based on incomplete information.
Global pipelines on the other hand need the access to the complete viewgraph and are not capable of delivering intermediate results.
In this paper we propose a new reconstruction pipeline working in a progressive manner rather than in a batch processing scheme.
The pipeline is able to recover from failed reconstructions in early stages, avoids to take binding decisions, delivers a progressive output and yet maintains the capabilities of existing pipelines.
We demonstrate and evaluate our method on diverse challenging public and dedicated datasets including those with highly symmetric structures and compare to the state of the art.
\end{abstract}

\section{Introduction}
3D reconstruction from individual photographs is a long studied topic in computer vision~\cite{Snavely2008,Agarwal2011,Heinly2015}.
The field of \gls{sfm} deals with the intrinsic and extrinsic calibration of sets of images and recovers a sparse 3D~structure of the scene at the same time.
Traditional methods are usually designed as batch processing algorithms, where image acquisition and image processing are separated into two independent steps.
This contrasts with current demand, when one would like to be able to convert an object or a scene into a 3D~model anytime and anywhere, just by using the mobile phone one is carrying in her pocket.
Recent developments in mobile technology and the availability of 3D~printers raised the need for 3D~content even more and underlay the importance of 3D~modeling.
In a user-centric scenario, images are taken on the spot and processed by a 3D~modeling pipeline on-the-fly~\cite{Kang2015,Locher2016}.
Any feedback which gets available to the user helps to guide her acquisition and, even more importantly, assures that the 3D~model represents the real-world object in the desired quality.
In a collaborative scenario, multiple users acquire pictures of the same object and images are gathered on the reconstruction server in the cloud.
The 3D~model is progressively built and intermediate reconstruction results are shown to the user.
Images are getting available as they are taken and the \gls{sfm} pipeline has never access to the complete set of information in the dataset.
Moreover, the whole reconstruction process might have a starting, but no predefined end point.
Users might always decide to add more images to an existing model.

\begin{figure}[t]
    \captionsetup[subfigure]{singlelinecheck=false, justification=centering}
    \centering
    \begin{subfigure}[b]{0.49\linewidth}
        \includegraphics[trim={8cm 5cm 8cm 3cm},clip,width=\textwidth]{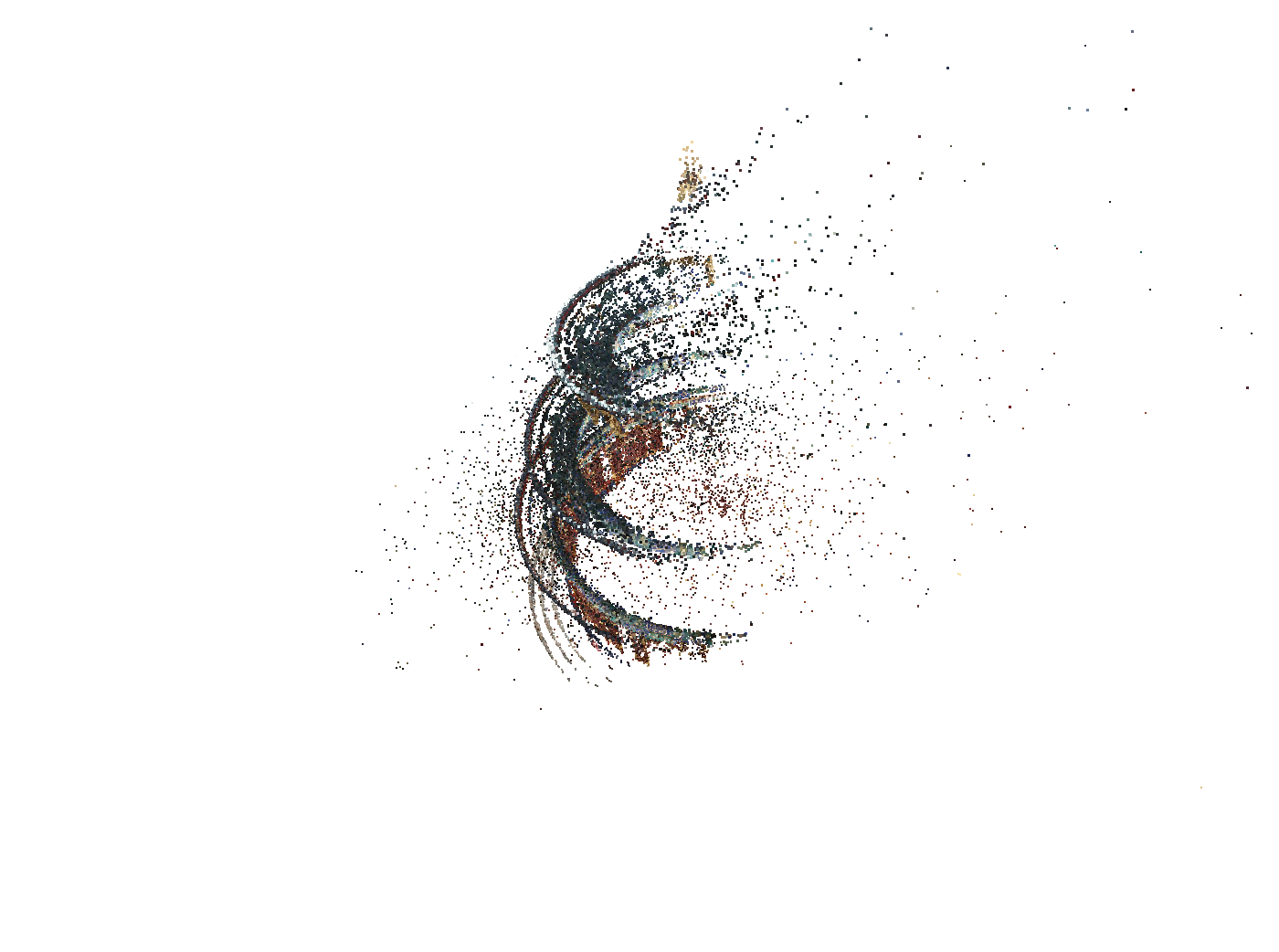}
        \caption{incremental SfM points}
        \label{fig:temple100-vsfm}
    \end{subfigure}
    \begin{subfigure}[b]{0.49\linewidth}
        \includegraphics[trim={10cm 6cm 10cm 6cm},clip, width=\textwidth]{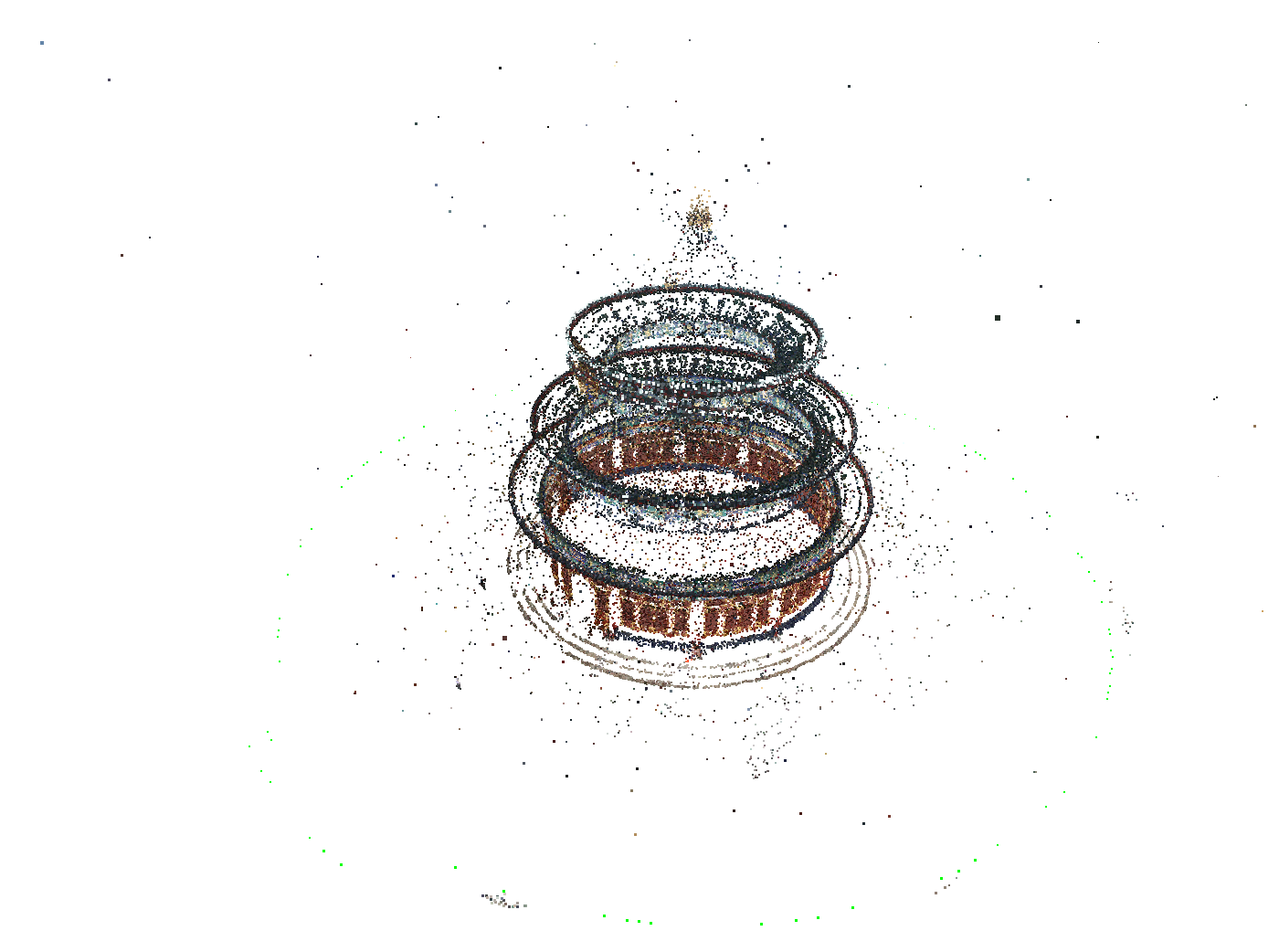}
        \caption{progressive SfM points}
        \label{fig:temple100-mysfm-points}
    \end{subfigure}
    \begin{subfigure}[b]{0.49\linewidth}
        \includegraphics[trim={10cm 6cm 10cm 6cm},clip,width=\textwidth]{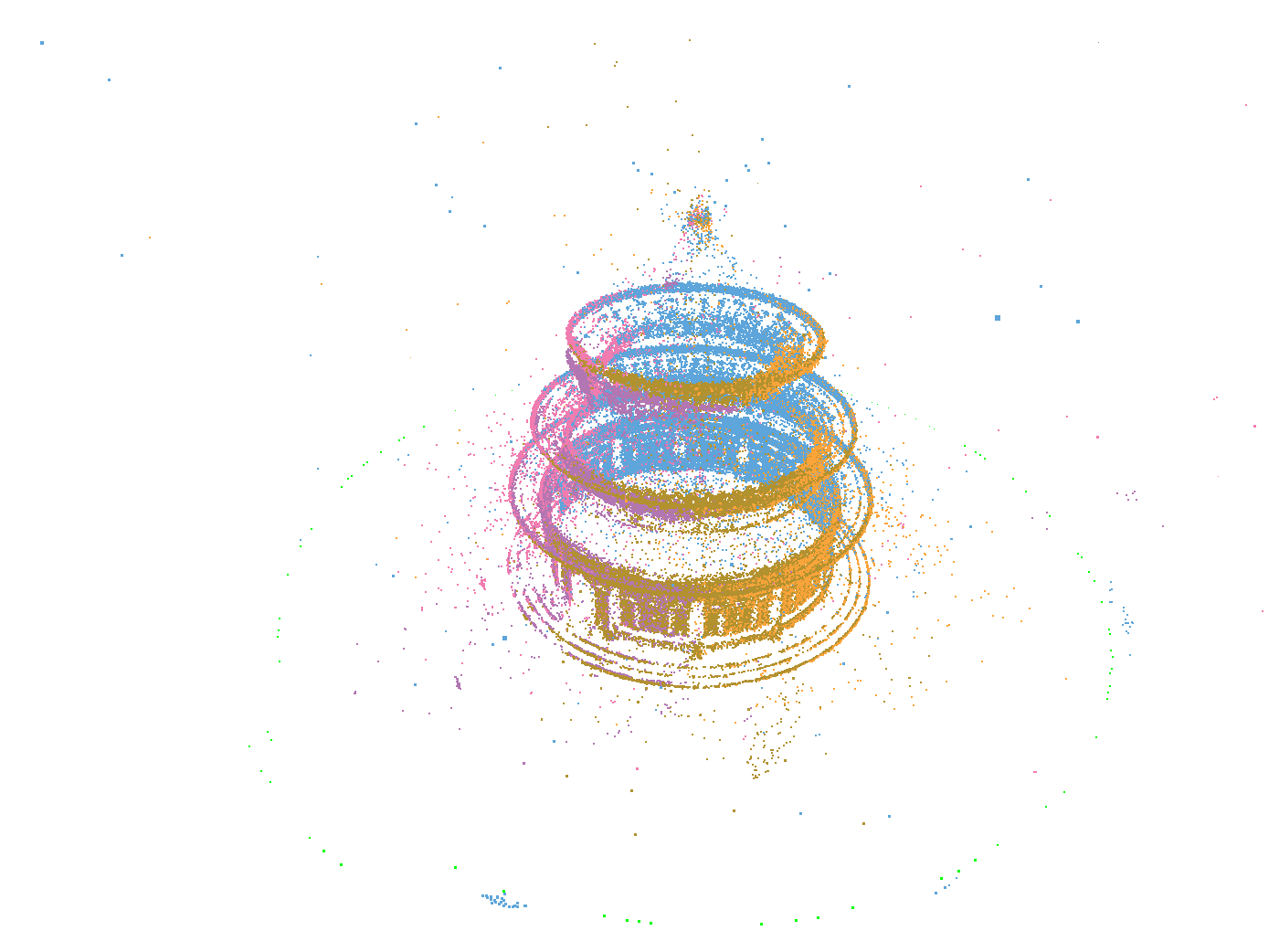}
        \caption{progressive SfM clusters}
        \label{fig:temple100-mysfm-cluster}
    \end{subfigure}
    \begin{subfigure}[b]{0.49\linewidth}
        \includegraphics[width=\textwidth]{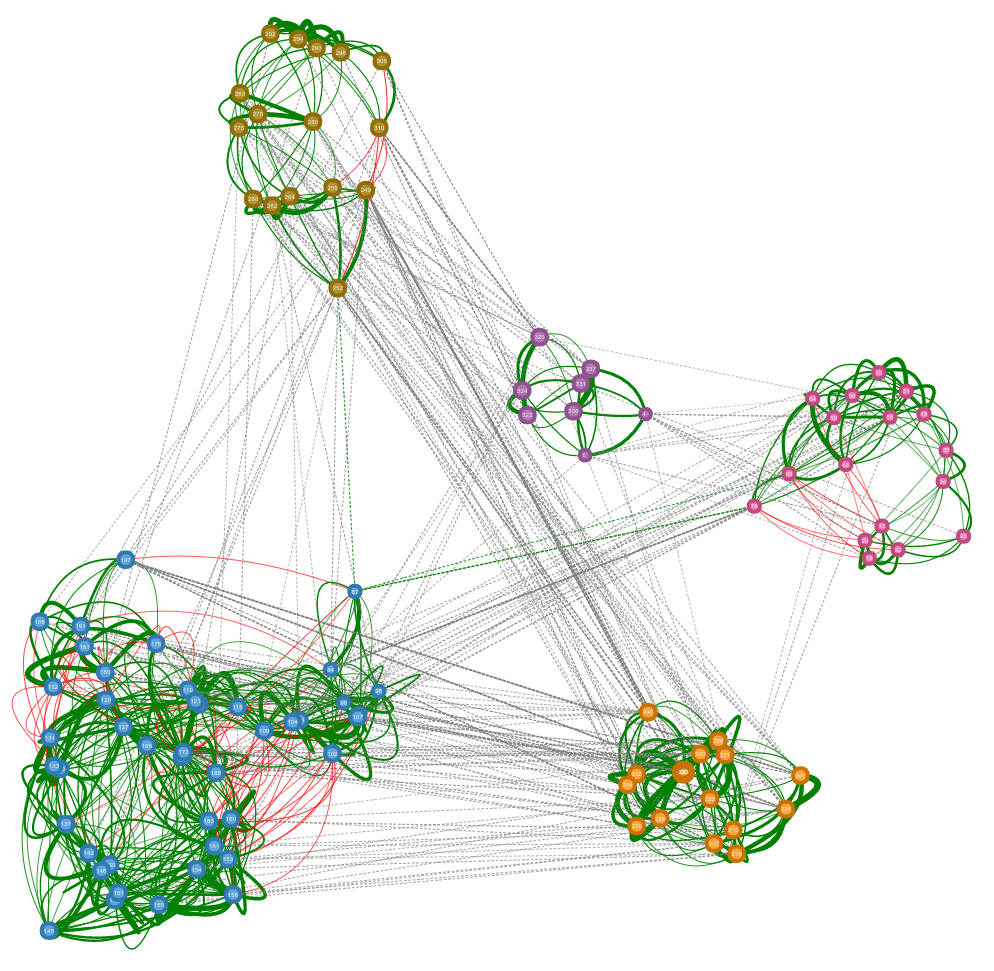}
        \caption{progressive SfM viewgraph}
        \label{fig:temple100-mysfm-viewgraph}
    \end{subfigure}
    \caption{Opposed to existing methods, a favorable image order is not crucial for the proposed progressive \gls{sfm} pipeline. While the baseline method (a) fails to recover the structure of the scene our method successfully reconstructs the temple (b). Individual clusters are identified in the viewgraph (d) and merged (c) based on a lightweight optimization.  }\label{fig:temple100}
\end{figure}

In this work we therefore propose a progressive \gls{sfm} pipeline which avoids taking (potentially fatal) binding decisions and therefore is as independent of the input image order as possible.
Moreover, the proposed pipeline reuses already computed intermediate results in later steps and is suited for delivering progressive modeling results back to the user within seconds.

\subsection{Related Work}

Classical \gls{sfm} pipelines are typically not suited to be used in such a progressive -- multiuser-centric scenario.
Global pipelines~\cite{Moulon2013,Sinha2012} start by estimating poses of all cameras in the dataset and estimate the structure in a second step.
Evidently, this relies on the access to the complete dataset which contradicts the idea of progressive 3D~modeling.
Sequential \gls{sfm} pipelines, sometimes termed SLAM~\cite{Davison2003,Mur2015}, are inherently suitable for processing (potentially infinite) streams of images.
Nevertheless, the underlying assumption often is that images neighboring in the sequence are spatially close in the scene, which is easily violated when streams from multiple users are combined together.
Incremental \gls{sfm} pipelines~\cite{Snavely2008,Wu2013,Schonberger2016} build a 3D~model by initializing the structure from a small seed and gradually growing it by adding additional cameras.
This scheme is closer to the requested progressive scenario but, unfortunately, is strongly dependent on the order in which images are added to the model~\cite{Martinec2007,Schonberger2014}.
View selection algorithms~\cite{Sweeney2015} carefully determine the image order usually by employing the global matching information which is not available in the progressive case.
Hierarchical \gls{sfm} pipelines~\cite{Havlena2010,Farenzena2009} try to overcome the problem of improper seed selection by starting from several seed locations at the same time and eventually merging the partial 3D~models into a single global model in later stages.
Sweeney~et~al.~\cite{Sweeney2016} group multiple individual cameras and optimize them jointly as a distributed camera.
However all hierarchical methods require the knowledge of all images in advance.
Our proposed method is partially inspired by these approaches but in order to provide the progressive capability, the hierarchy is not fixed but rather re-defined every time new images are added to the reconstruction process.

Due to the lack of global information, an incremental pipeline would connect new images to the existing model based on incomplete information which often causes corrupted 3D~models~\cite{Zach2008,Roberts2011,Jiang2012,Heinly2014}.
Even more importantly, incremental pipelines cannot recover from wrong decisions taken based on missing information and therefore can easily get stuck in only locally optimal reconstructions.
The only reliable solution would be re-running an incremental or global pipeline from scratch when new images become available which leads to an impractical algorithm runtime.
Heinly et al.~\cite{Heinly2014} detected erroneous reconstructions of scenes with duplicate structure in a post processing step by evaluating different splits of the model into submodels and potentially merge them in the correct configuration by leveraging conflicting observations.
Our method in contrary is a full fledged \gls{sfm} pipeline which avoids getting trapped in a local minima in the first place.
Faulty configurations are detected and corrected on-the-fly and not in a post processing step.

Our work bases on a lightweight representation of the complete scene as a viewgraph.
Many existing approaches investigated robustification of the global view-graph by filtering out bad epipolar geometries~\cite{Sweeney2015,Cui2015} and enforcing loop constraints~\cite{Zach2010}.
Wilson and Snavely~\cite{Wilson2014} are able to reconstruct scenes with repetitive structures by scoring repetitive features using local clustering.
Recent work of Cui~\cite{Cui2017} takes a similar approach to our pipeline.
The Hybrid \gls{sfm} pipeline estimates all rotations of the viewgraph in a community based global algorithm.
The estimated orientations are leveraged in the second phase of the pipeline by estimating the translations and structure of the scene in an incremental scheme with reduced dimensionality.
While sharing the idea of combining global and incremental schemes, HSfM is a pure batch processing algorithm.
The global rotation averaging needs access to the complete view graph in advance which is not available in a progressive scheme.
In our work we combine a dynamic global view graph with a local clustering based on a connectivity score and combine the advantages of incremental and global structure from motion.
In order to accommodate for the demands of a progressive pipeline, we allow for flexibility in already reconstructed parts of the model and constantly verify the local reconstructions against the globally optimized viewgraph.

\subsection{Contributions}
We propose a novel progressive \gls{sfm} pipeline which enables 3D~reconstruction in an anytime anywhere multiuser-centric scenario.
Unlike traditional pipelines, the proposed approach avoids taking binding decisions, does not depend on the order of incoming images and is able to recover from wrong decisions taken due to the lack of information.
Moreover, the computed intermediate results are propagated along the reconstruction process resulting in an efficient use of computational resources.

\section{Progressive Reconstruction}

In the following section we give an overview of our progressive \gls{sfm} pipeline and detail individual components and key aspects later on.

\subsection{Overview}

\begin{figure}
  \centering
  \includegraphics[width=0.85\linewidth]{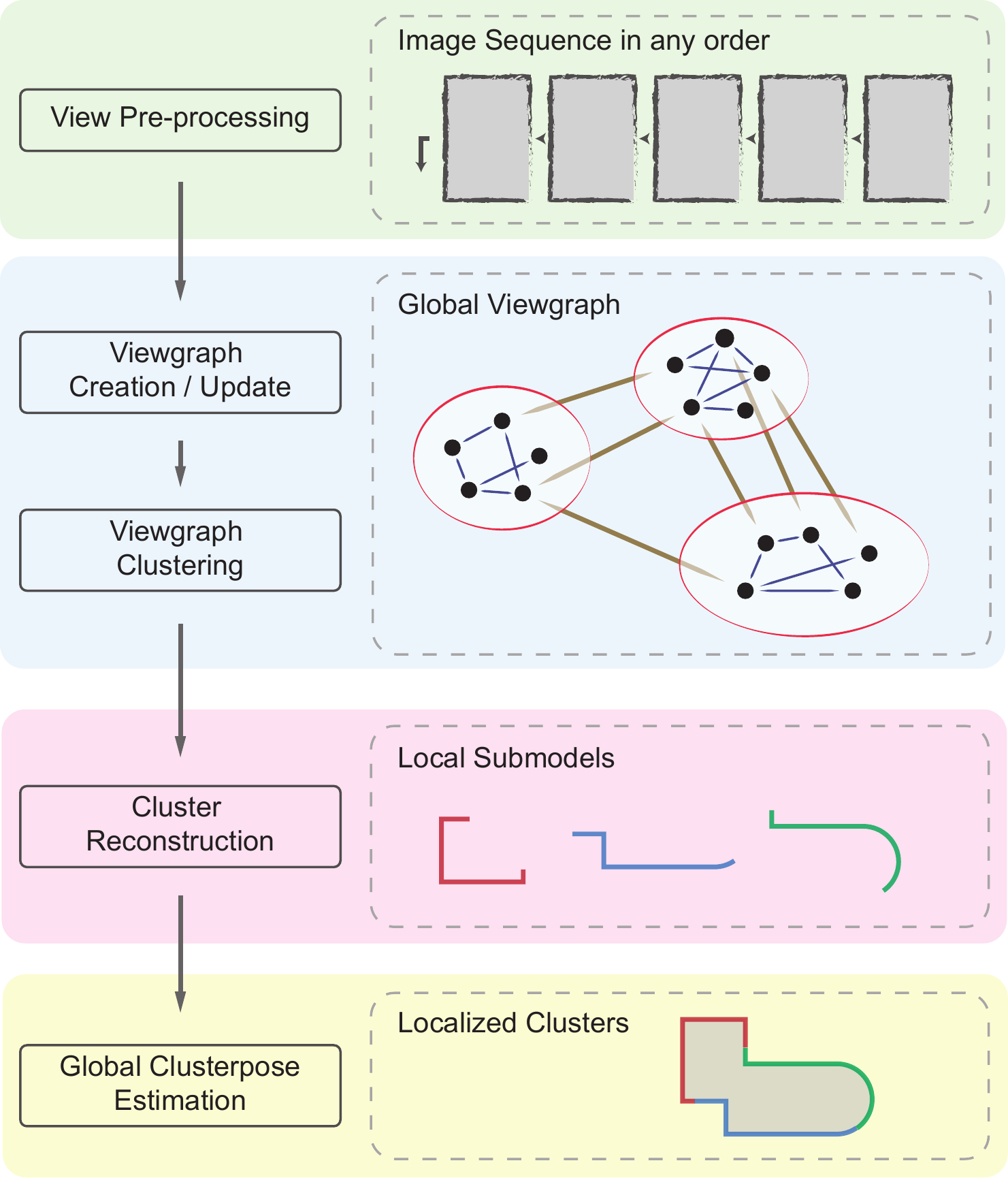}
  \caption{
    An overview of the main steps of the progressive \gls{sfm} pipeline and its involved components.
  }
  \label{fig:overview}
\end{figure}

Our progressive \gls{sfm} pipeline takes an ordered sequence of images with its geometrically verified correspondences as the input and delivers a sparse pointcloud and calibrated camera poses as the output (see Figure~\ref{fig:overview}).
The resulting sparse configuration is updated with every image added to the scene.
A viewgraph with nodes being images and edges connecting pairs of matched images is gradually built and serves as the global knowledge throughout the whole reconstruction.
On every iteration of the algorithm, the viewgraph is clustered based on local connectivity and individual clusters are processed locally.
In each of the local clusters a robust rotation averaging scheme filters out wrong two-view geometries and the 3D~structure is estimated using either an incremental or a global \gls{sfm} pipeline.
The cluster configuration between two time steps is tracked and the already estimated parts of the 3D~model are passed to the next stage.
The global configuration of individual clusters is estimated in the last stage by robustly estimating 7~DoF similarity transforms between them using the remaining inter-cluster constraints.
Generally, the local incremental method enables robust and efficient reconstructions while the viewgraph combined with the robust rotation averaging injects the global knowledge and allows for correction of corrupted 3D~models.

\subsection{Progressive Viewgraph}
The algorithm takes a (randomly) ordered sequence of images $\mathbf{I} = \left( I_0, I_1, I_2, \dots \right)$ as the input.
Every incoming image is matched against the most relevant images already present in the scene and geometrically verified pairwise correspondences are obtained.
A viewgraph $\mathcal{G}\left(\mathcal{V}_t, \mathcal{E}_t \right)$ with images as vertices $\mathcal{V}_t = \left\lbrace I_i \vert i \leq t \right\rbrace$ is maintained at every time step $t$.
Two vertices $( V_i, V_j )$ are connected by an undirected edge $\mathcal{E}_{ij}$ iff there exists a minimum amount of correspondences $\left( \mathbf{M}_{ij} > \eta \, \wedge \, i < j \leq t \right)$ between them where $\mathbf{M} \in \mathbb{R}^{t \times t}$ is the matching matrix.
Every edge $\mathcal{E}_{ij}$ has an associated relative rotation $\mathbf{R}_{ij}$ and translation direction $\mathfrak{t}_{ij}$ which is obtained by decomposing either the estimated essential matrix in the calibrated case or the fundamental matrix when the focal length is not known.

The order in which images are fed to a reconstruction pipeline plays an important role and every snapshot of the viewgraph only captures the past information of the reconstruction process.
This is why filtering of supposedly wrong two-view geometries at this stage can be very dangerous.
It might happen that a geometry is inconsistent with other local geometries in the neighborhood\footnote{This can, e.g.,\ be checked by computing the cumulative rotation of loops in which the edge is participating~\cite{Zach2010}} at the current time-step but connecting images added in later steps may show that the two-view geometry was actually correct.
As a wrong decision in the global viewgraph could lead to a local minimum in the reconstruction, i.e.\ a corrupted 3D~model, we do not conduct any outlier rejection and defer robustification to a later stage.

\subsection{Clustering}
Motivated by the general observation that densely connected regions of the viewgraph are likely to form a 3D~model worth reconstruction, the viewgraph is clustered in a second step.
The distance $d_{ij}$ between two vertices $V_{ij}$ is based on the weighted Jaccard distance of the adjacent edges where connections to neighboring vertices are weighted by the number of verified correspondences.

\begin{equation}
d_{ij} = 1 - \left. \dfrac{\sum\limits_{n\in \mathcal{N}} \mathbf{M}_{in} +  \mathbf{M}_{nj}}{\sum\limits_{n=0}^{t} \mathbf{M}_{in} + \mathbf{M}_{nj}} \right\vert \mathcal{N} = \left\lbrace k \vert \mathbf{M}_{ik} \cdot \mathbf{M}_{kj} > 0 \right\rbrace
\end{equation}

A set of clusters $\mathcal{C}_t$ is hierarchically grown by single-linkage clustering until no single edge with $d_{ij} < \eta$ between two clusters exists.
Single-linkage clustering tends to generate clusters with a chain-like topology where the two ending nodes might have a distance way larger than the defined threshold.
While this might be a disadvantage in other applications it is actually beneficial in our application as local (incremental) reconstruction pipeline performs well for such graph structures.

\subsubsection*{Incremental Clustering}
Due to the single-linkage hierarchical clustering, a simplified incremental scheme can be used to update an existing cluster topology with a new node.
While generally a single extra node can cause a complete change in topology of the clustered graph, the changes are limited to clusters which are connected to the new node.
As a result only clusters with an edge connecting to the new node have to be re-clustered which leads to an efficient and scalable implementation.
Note that in worst case the whole graph still might be updated -- but in most cases a new image only connects to few clusters and therefore most of the existing clusters remain untouched.

\subsection{Cluster Tracking and Recycling}
Between the transition of two timesteps $t_b$ and $t_a = t_b + 1$ the topology of clusters can undergo large changes but mostly will either stay the same or be extended by the new image.
All changes in the clusters have to be propagated to the eventually estimated 3D~structure.
We therefore keep track of the nodes changing cluster between the two timesteps and add or remove the corresponding images in the local reconstruction.
The recycling of intermediate (partial) 3D reconstructions can be realized by a merge and split scheme.
If a group of interconnected nodes transfer together from one cluster to another, the corresponding cameras in the 3D model can be separated from the rest and merged into the potentially existing structure of the new cluster.

\subsection{Cluster Reconstruction}
Once individual clusters $\mathcal{C}_i$ have been identified by the hierarchical clustering, the 3D~structure of the images and correspondences of every cluster $C$ is estimated.
The cluster reconstruction process is only triggered if its topology has changed meaning either some nodes were added or some nodes were removed from the cluster.
If the topology of the cluster is unchanged between the two time steps $t_b$ and $t_a$, the reconstruction step is skipped as a whole.

A sub-graph $\mathcal{V}_C$ capturing all vertices and edges of the cluster $C$ is extracted in a first step.
All following operations are restricted to the scope of the extracted graph $\mathcal{V}_C$.

\vspace{-0.3em}
\subsubsection*{Robust Rotation Averaging}
The unfiltered viewgraph is potentially corrupted by outliers and has to be cleaned up for further usage.
As this step is performed in every iteration it is safe to reject outlier two-view geometries.
As a result of the repetitive execution of the filtering stage, it is important to use a computationally efficient filtering scheme.
As proposed by Chatterjee et al.~\cite{Chatterjee2013} we estimate the global rotations of the subgraph with a robust $\ell_1$ optimization.
Global Rotations $\mathbf{\bar{R}}$ are initialized by concatenating the relative rotations of a \gls{mst} extracted from the viewgraph $\mathcal{V}_C$ using the  number of verified inliers $\mathbf{M}_{ij}$ as edge weights.
The global rotations are then optimized by minimizing the relative rotation errors $\rho_{ij}$.

\begin{equation}
\arg \min\limits_{\mathbf{\bar{R}}} \sum\limits_{(i,j)\in \mathcal{E}_C} \rho_{ij} \left( \mathbf{R}_{ij}, \mathbf{\bar{R}}_j \mathbf{\bar{R}}_i^{-1} \right)
\end{equation}

By using the Lie-algebraic approximation of the relative rotation, the error can be expressed as a difference of the corresponding rotations $\mathbf{\omega}$. Where $\mathbf{\omega} = \mathit{\Theta} \mathbf{n} \in \mathfrak{so}(3)$ denotes a rotation by angle $\mathit{\Theta}$ around the unit axis $\mathbf{n}$.

\begin{equation}
\mathbf{\omega}_{ij} \approx \mathbf{\omega}_j - \mathbf{\omega}_i
\end{equation}

\noindent The relative rotations can be encoded in a sparse matrix $\mathbf{A}$ where each row only has two nonzero ${-1,+1}$ entries.
The robust $\ell_1$ norm combined with an edge weighting $\rho$ depending on the number of verified correspondences allows us to obtain the global rotations of the cluster $C$ in an iterative scheme by optimizing Eqn.~\ref{eq:minimizer} in every step.
For more details we refer the reader to the original publication~\cite{Chatterjee2013}.

\begin{equation}
\label{eq:minimizer}
\arg \min\limits_{\mathbf{\omega}_g} \left\Vert \mathbf{A} \, \Delta \mathbf{\omega}_{g} - \mathbf{\rho} \mathbf{\omega}_{rel}  \right\Vert_{\ell_1}
\end{equation}

\noindent The obtained global rotations can optionally be refined by \gls{irls} method using a robustified $\ell_2$ norm.
As we are primarily interested in rejecting outliers, the results of the $\ell_1$ optimization is accurate enough and we omit the refinement step.

Finally edges with a relative error above $\rho_{g \max}$ are removed from the local viewgraph.

\vspace{-0.3em}
\subsubsection*{Reconstruction}
\begin{figure}
  \centering
  \includegraphics[width=0.85\linewidth]{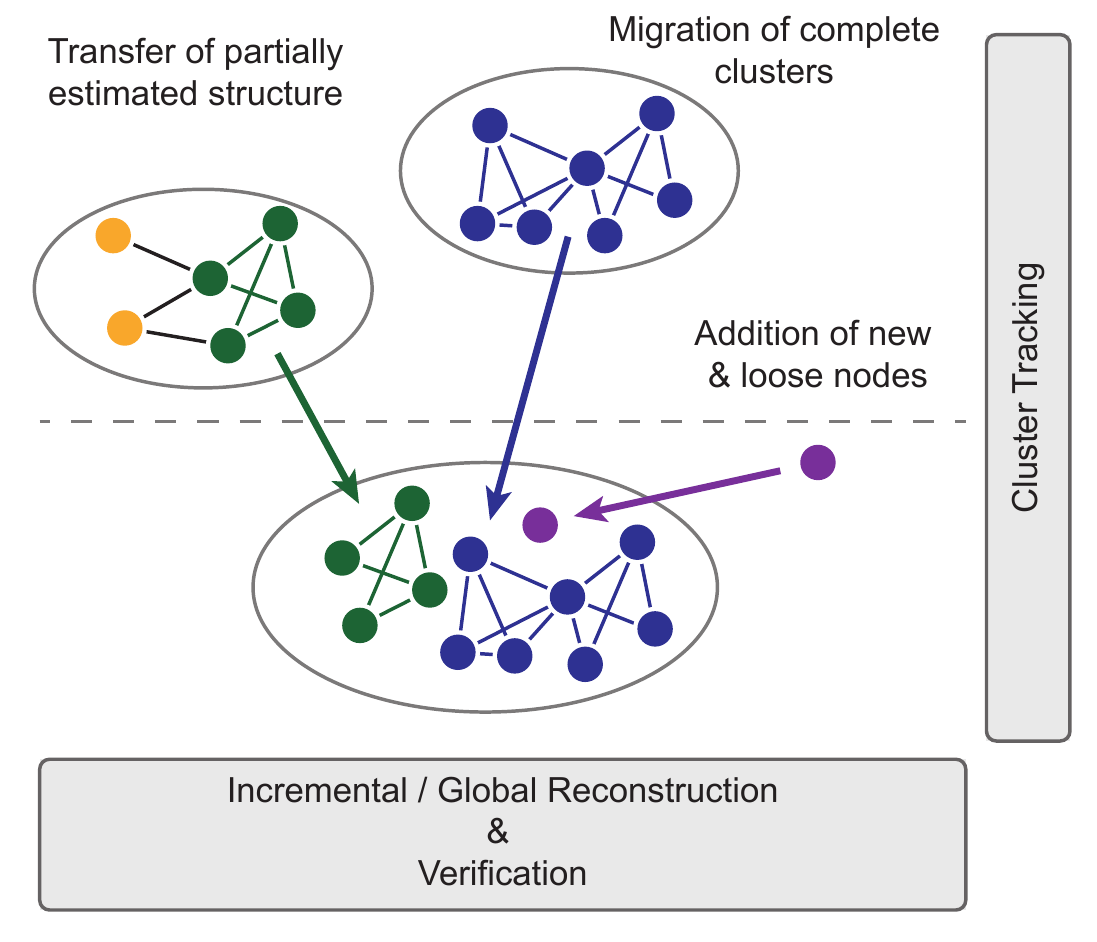}
  \caption{
    Depending on the status of an individual cluster, its structure is estimated with different methods.
  }
  \label{fig:reconstruction}
\end{figure}

Using the global rotations and the pairwise matches as the input, the structure of the individual clusters is estimated with different methods depending on the current status (Figure~\ref{fig:reconstruction}).

\begin{description}

\item [New:]
If we could not recover any structure from $t_b$ and the cluster contains at least $\mu_{min}$ nodes but less than $\mu_{max}$ images a new reconstruction is starting using an incremental \gls{sfm} pipeline similar to~\cite{Snavely2008,Wu2013}.
If the number of images in the cluster is larger than $\mu_{max}$, incremental pipelines get inefficient and a global pipeline is more suited.
We therefore refine the global rotations and estimate global position using the 1DSfM method~\cite{Wilson2014}. The structure is then obtained by triangulating consistent feature-tracks and the configuration is refined by a bundle adjustment step.

\item [Extend:]
In most cases an existing local reconstruction can be extended by one or multiple images which are either new or transferred from another cluster.
In order to extend an existing 3D reconstruction with a new image we extend existing feature tracks by the new correspondences and estimate the 3D position of the camera by the P3P algorithm~\cite{Kneip2011}.
New tracks are added and potential conflicting tracks are split up.
A local bundle adjustment refines the structure and calibration of the newly added cameras.
If the model has grown by more than $\eta_{grow}$ percent, a bundle adjustment step over the local reconstruction refines the whole structure.

\item [Transfer:]
If a large part (more than $\mu_{min}$ nodes) of an already estimated local model is transferred to another cluster, its already estimated structure is kept and transferred to the new cluster.
Commonly estimated tracks are fused and a bundle adjustment step refines the structure.
Potentially unestimated cameras are then added in an incremental scheme as described before.

\end{description}

\vspace{-0.3em}
\subsubsection*{Detection of Bad Configurations}
Due to the incremental reconstruction scheme for cluster reconstruction we can reuse most of the intermediate results from an earlier stage and propagate them to later stages.
The incremental scheme is highly dependent on the actual image ordering and therefore some unfortunate decisions taken in early stages (e.g.\ wrongly connecting an image due to missing information) cannot be recovered in later stages.
Global clustering usually solves this problem for us as the wrongly connected image or sub-model is likely to be transferred to another cluster at a later stage.
But it might also happen that the image actually belongs to the same cluster and yet is wrongly connected.
Without any additional counter measurements, we would end up with a corrupted reconstruction in such cases.

By using the relative global rotations  of the local viewgraph (which is independent of the image order) we can evaluate an error measure between the rotations of the current local reconstruction $\mathbf{\hat{R}}$ and the global rotations $\mathbf{\bar{R}}$.

\begin{equation}
\rho_{l} = \left\Vert \, \left\lbrace \left. \mathbf{\bar{R}}_{ij} \, \mathbf{\hat{R}}_{ij}^{-1} \right\vert (i,j) \in \mathcal{E}_C \right\rbrace \, \right\Vert_{\infty}
\end{equation}

\noindent If the rotation error $\rho_{l}$  is above a certain limit $\rho_{l\max}$ the reconstruction is likely to be erroneous.
Hence it is reset and re-estimated in the next phase of the reconstruction.

\subsection{Cluster Pose Estimation}
The output of the pipeline so far are multiple locally highly consistent but disconnected model parts.
In order to merge them to a final representation we make use of the remaining inter-cluster connections.
As this step of merging multiple models is often rather fragile and can easily go wrong, several methods employed a rather conservative merging criteria as multiple cameras of overlap~\cite{Havlena2010} or a high amount of common points~\cite{Wu2013}.
Instead, we are estimating pairwise similarity transforms $\mathcal{T}_{ab} \in \mathit{Sim}(3)$ between clusters $C_a$ and $C_b$ and optimize the global positions of the cluster centers.
For robustification, we use a two stage verification scheme for individual inter-cluster constraints.

In the first stage, a similarity transform based on 3D-to-3D correspondences from every single edge $\mathcal{E}_{ij}$ between clusters $C_a$ and $C_b$ is estimated in a RANSAC scheme~\cite{Horn1987,Fischler1981}.
The vertices (camera centers) of $C_a$ are afterwards transformed using the estimated transformation and the camera configuration of the obtained combined model is compared to the individual configuration before.
A neighborhood similarity measure $\mathfrak{s}$ evaluates how many of the cameras $\mathbf{p}_k$ in the merged cluster would retain their closest neighbors $\mathrm{NN}_a(\mathbf{p}_k)$ and $\mathrm{NN}_b(\mathbf{p}_k)$ for the cameras from clusters $C_a$ and $C_b$, respectively. $\mathbf{p}_k$ denotes the camera center of the node $k$ and the operator $\mathcal{T}_{ij} \circ \mathbf{p}_k$ transforms a point by the similarity transform.

\begin{equation}
  \mathcal{T}_{ij} \; \forall \; i \in C_a \wedge j \in C_b \wedge (i,j) \in \mathcal{E}_C
\end{equation}

\begin{align}
  \mathbf{P}_a &= { \mathbf{p}_k \vert k \in C_a } \\
  \mathbf{P}_b &= { \mathbf{p}_k \vert k \in C_b } \\
  \mathbf{P}_c &= \mathbf{P}_b \cup {\mathcal{T}_{ij} \circ \mathbf{p}_k \vert k \in C_a }
\end{align}
\begin{equation}
  \mathfrak{s} = \dfrac{1}{\vert \mathbf{P}_c \vert} \sum\limits_{\mathbf{p}_k \in \mathbf{P}_c} s_k
\end{equation}
\begin{equation}
s_k =   \begin{cases}
1 & k \in C_a \wedge \mathcal{T}_{ij} \circ \mathrm{NN}_a(\mathbf{p}_k) = \mathrm{NN}_c(\mathcal{T}_{ij} \circ \mathbf{p}_k) \\
1 & k \in C_b \wedge \mathrm{NN}_b(\mathbf{p}_k) = \mathrm{NN}_c(\mathbf{p}_k) \\
0 & \mathrm{otherwise}
\end{cases}
\end{equation}

\noindent Edges with the neighborhood similarity measure below $\lambda_c$ are rejected as outliers for the final constraint estimation.
The measure is motivated by the observation that nearby cameras should already be clustered by the original algorithm and prevents merged models where the cameras are extensively interleaved.

In the second stage, a final transformation $\mathcal{T}_{ij}$ between the clusters is estimated using all correspondences of edges that survived the first filtering stage.
All relative constraints are gathered and a cluster graph $\mathcal{G}_C$ with local reconstructions as nodes and pairwise inter-cluster constraints as edges is created.

\begin{equation}
\arg \min\limits_{\mathcal{P_C}} \sum\limits_{(a,b)\in \mathcal{G}_C} \left\Vert \delta\left( \mathcal{T}_{ab} \, \mathcal{P}_b \mathcal{P}_a^{-1} \right)\right\Vert ^2
\end{equation}

\noindent The global poses $\mathcal{P_C}$ of the clusters are afterwards obtained by iteratively minimizing the pairwise relative pose constraints $\mathcal{T}_{ab}$, where $\delta(.)$ denotes the robust Huber error function.

While this in the optimal case leads to a single 3D~model, the flexibility and efficiency of the global optimization allows us to choose a much less conservative threshold for cluster merging.
In addition to its efficiency, the cluster graph representation can easily incorporate additional extrinsic constraints, e.g.,\ the gravity vector for orientation with respect to the ground plane or coarse cluster-wise GPS locations for geo-referencing.
Such additional constraints can help to put individual models in relative perspective even if pure vision based inter-cluster constraints are not (yet) available.

\section{Experiments and Results}

In the following section, we first give some implementation details of the proposed solution and introduce the conducted experiments.

\subsection{Implementation}
We implemented the proposed pipeline as a C++ software.
SIFT~\cite{Lowe2004} features are detected on all images and described with the RootSift descriptor~\cite{Arandjelovic2012}.
The pipeline is setup for stream processing dedicated by the progressive reconstruction scheme.
Every incoming image is indexed by a 1M word vocabulary trained on the independent Oxford5k~\cite{Philbin2007} dataset and the top 100 nearest neighbors are retrieved using a fast tf-idf scoring~\cite{Sivic2003}.
Matches are geometrically verified by estimating the fundamental or essential matrix (depending on the availability of the focal length) as well as a homography matrix in a RANSAC scheme using the general USAC framework~\cite{Raguram2013}.
The relative rotation and translation direction are obtained by decomposing the fundamental, essential, or homography matrix (depending on the amount of inliers).
The incremental and global reconstruction of cluster centers bases on the publicly available Theia library~\cite{theia-manual} and the final clustergraph optimization is realized in the g2o~\cite{Kummerle2011} framework.
The following parameters were used within all the conducted experiments:
$\mu_{min} = 5$, $\mu_{max} = 50$, $\lambda_s = 0.9$, $\rho_{l\max} = \rho_{g\max} = 10^{\circ}$, $\eta = 0.2$, $\eta_{grow} = 0.15$.

\subsection{Baselines}
Throughout our expermiments we compare the proposed progressive pipeline against two linear pipelines (VisualSFM~\cite{Wu2013} and incremental Theia~\cite{theia-manual}) as well as to the recently published
hybrid pipeline HSfM~\cite{Cui2017}.
For an unbiased comparision, all geometrically verified matches were precomputed and imported into the various pipelines.
The streaming part of the pipeline was skipped and matches directly loaded from disk.

\begin{figure*}[t]
    \centering
    \begin{subfigure}[b]{0.24\textwidth}
        \includegraphics[width=\textwidth]{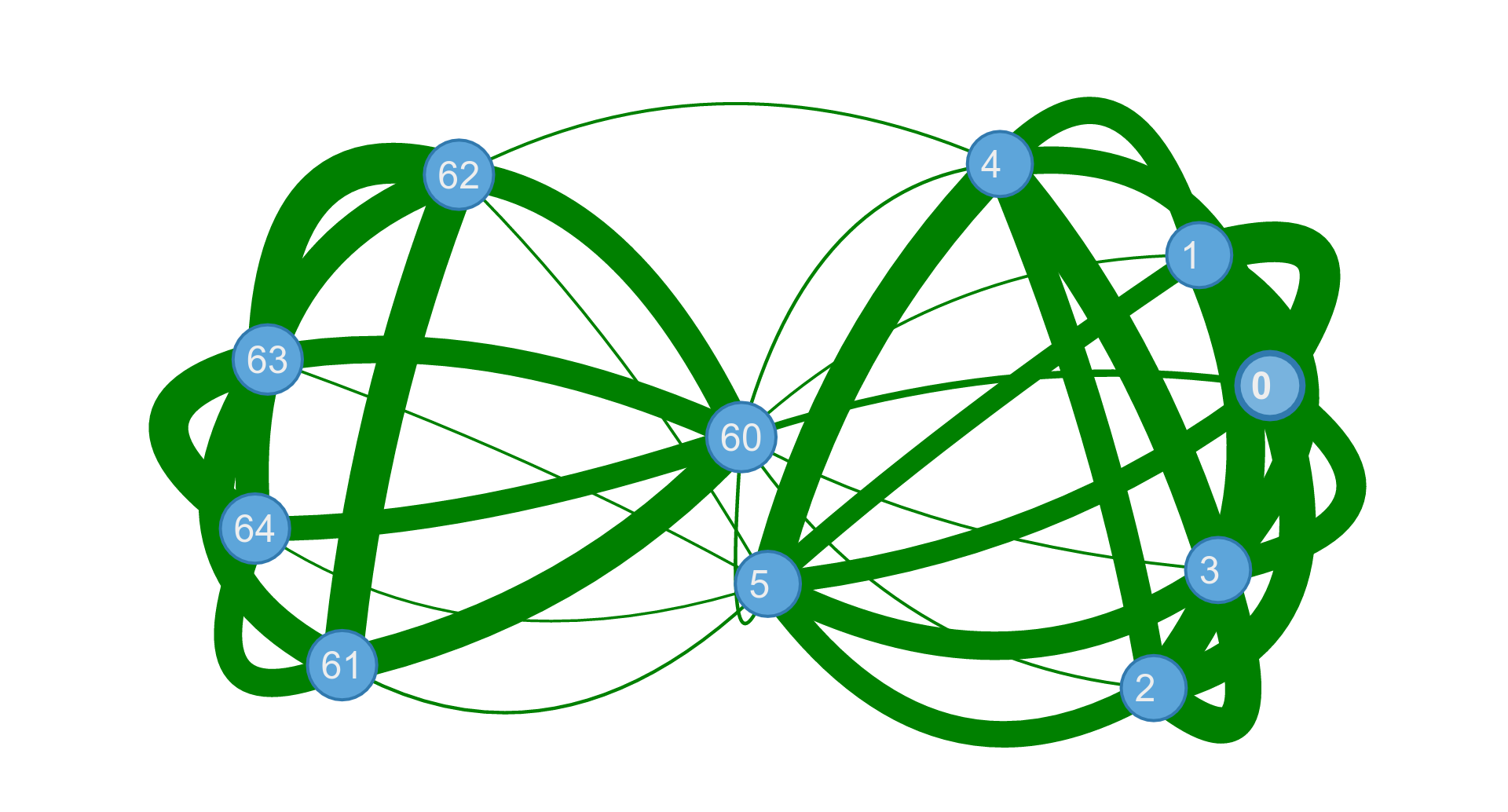}
        \includegraphics[trim={10cm 6cm 5cm 2cm},clip,width=\textwidth]{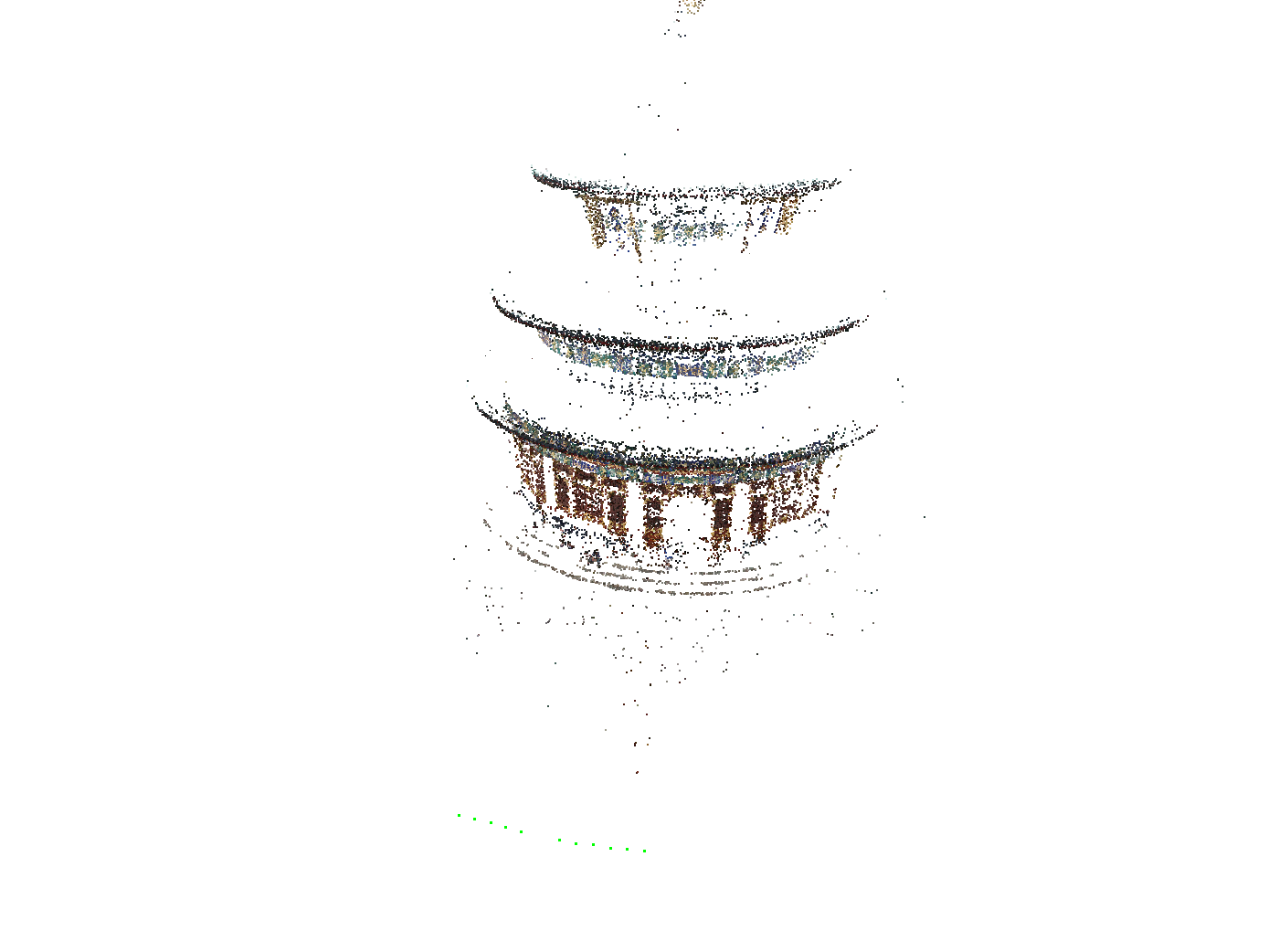}
        \caption{$t = 10$}
        \label{fig:temple60-10}
    \end{subfigure}
    \begin{subfigure}[b]{0.24\textwidth}
         \includegraphics[width=\textwidth]{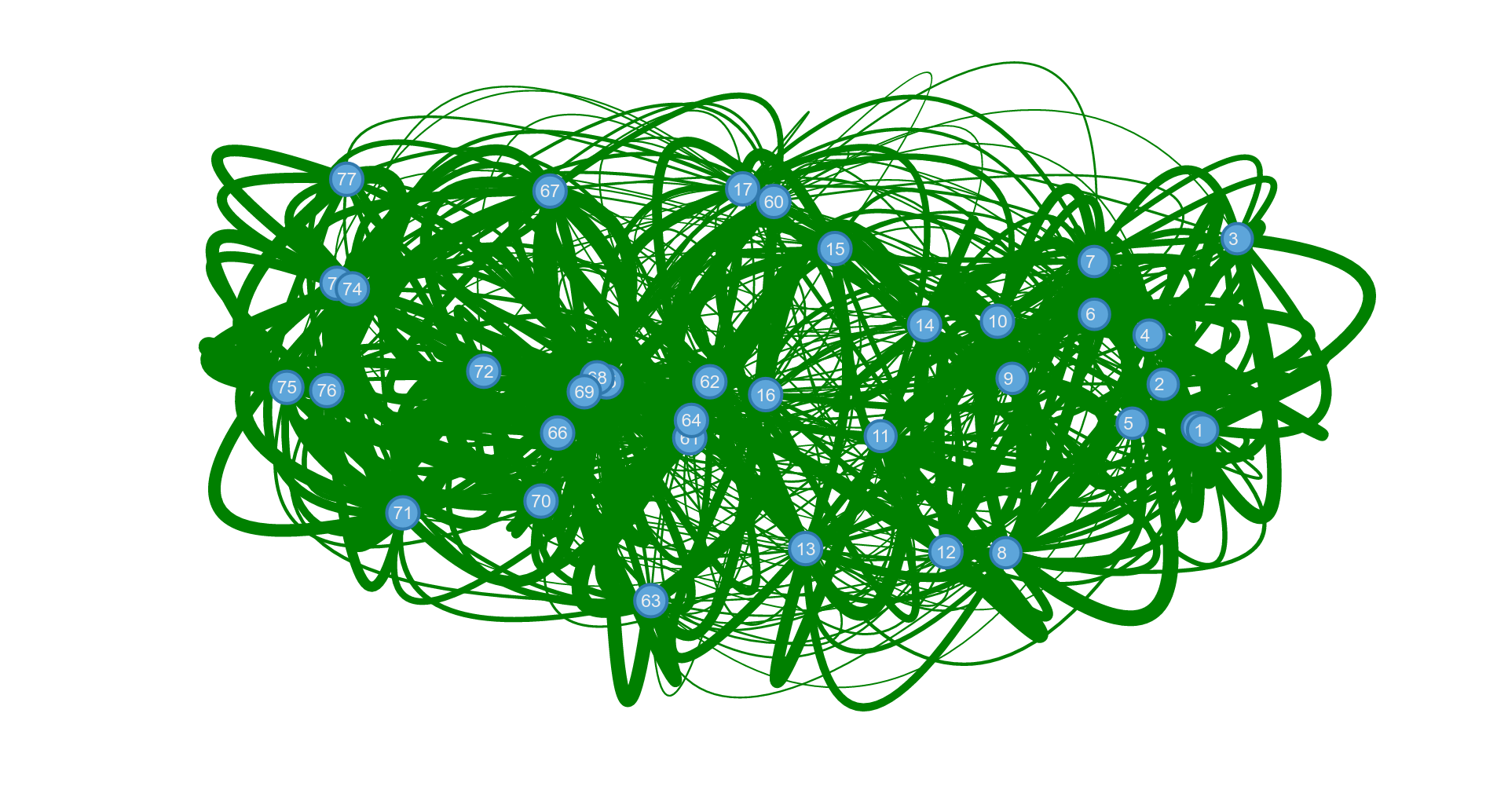}
         \includegraphics[trim={10cm 6cm 5cm 2cm},clip,width=\textwidth]{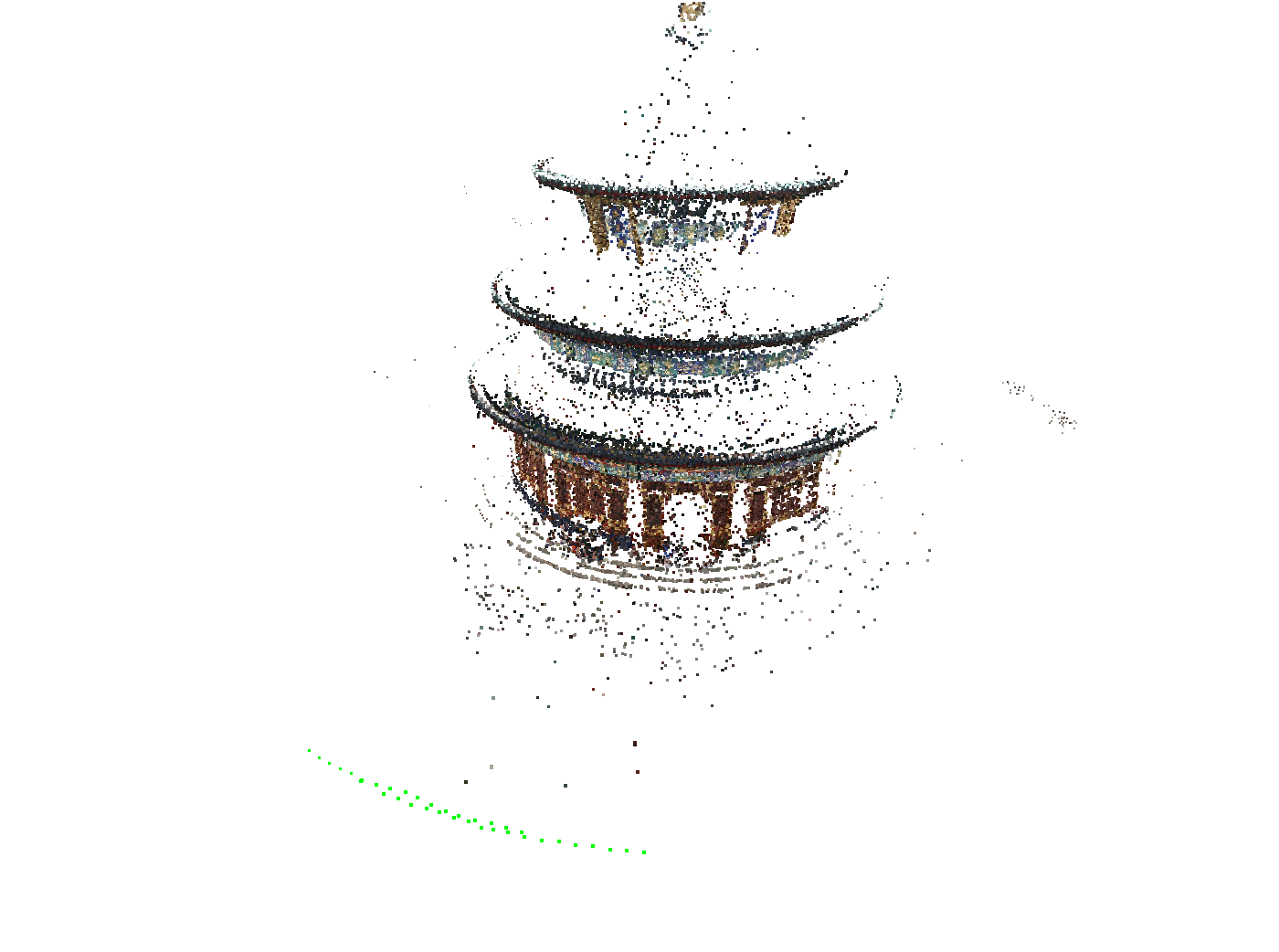}
        \caption{$t = 35$}
        \label{fig:temple60-35}
    \end{subfigure}
    \begin{subfigure}[b]{0.24\textwidth}
         \includegraphics[width=\textwidth]{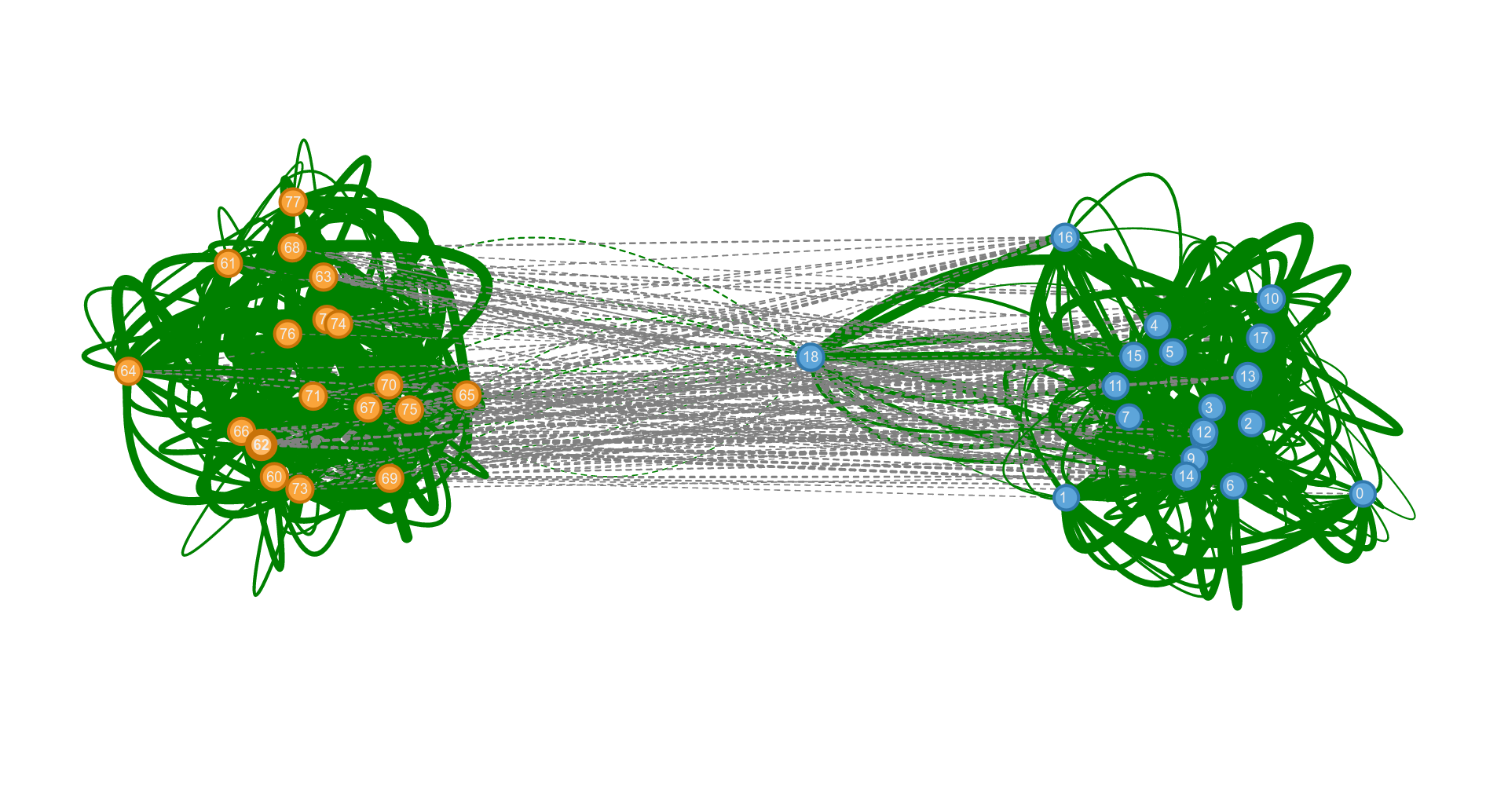}
         \includegraphics[trim={10cm 6cm 5cm 2cm},clip,width=\textwidth]{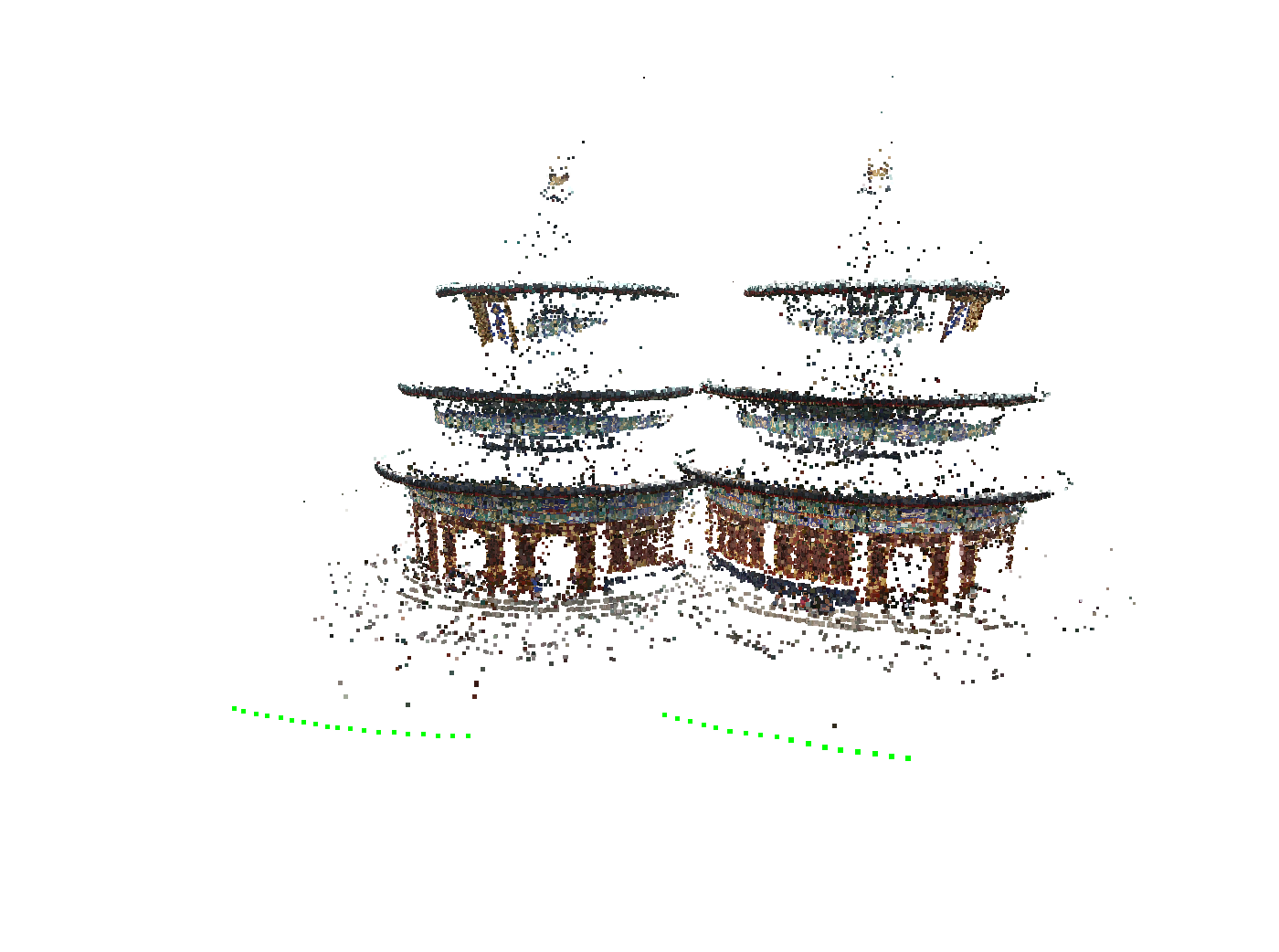}
        \caption{$t = 36$}
        \label{fig:temple60-36}
    \end{subfigure}
    \begin{subfigure}[b]{0.24\textwidth}
         \includegraphics[width=\textwidth]{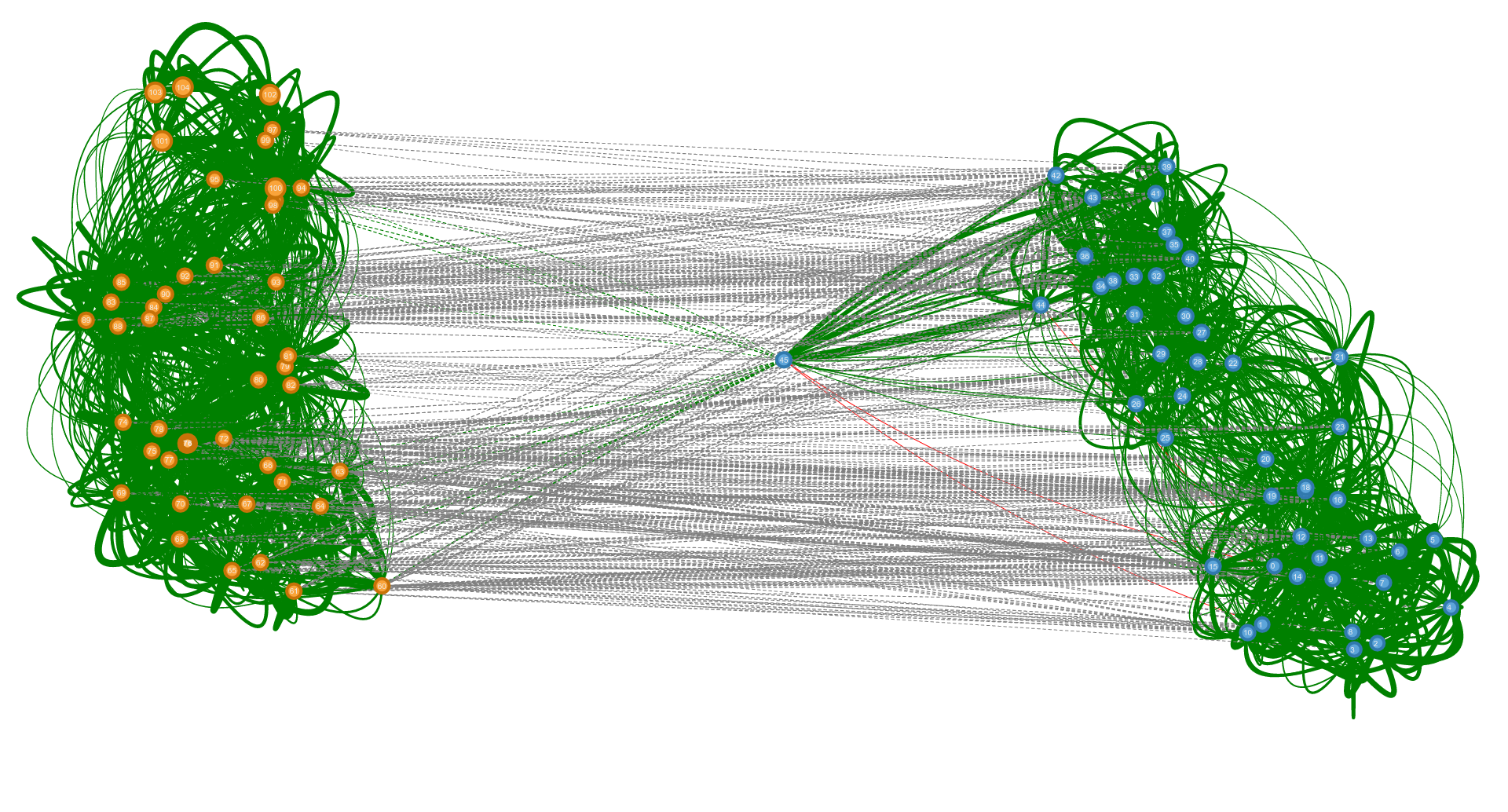}
         \includegraphics[trim={10cm 6cm 5cm 2cm},clip,width=\textwidth]{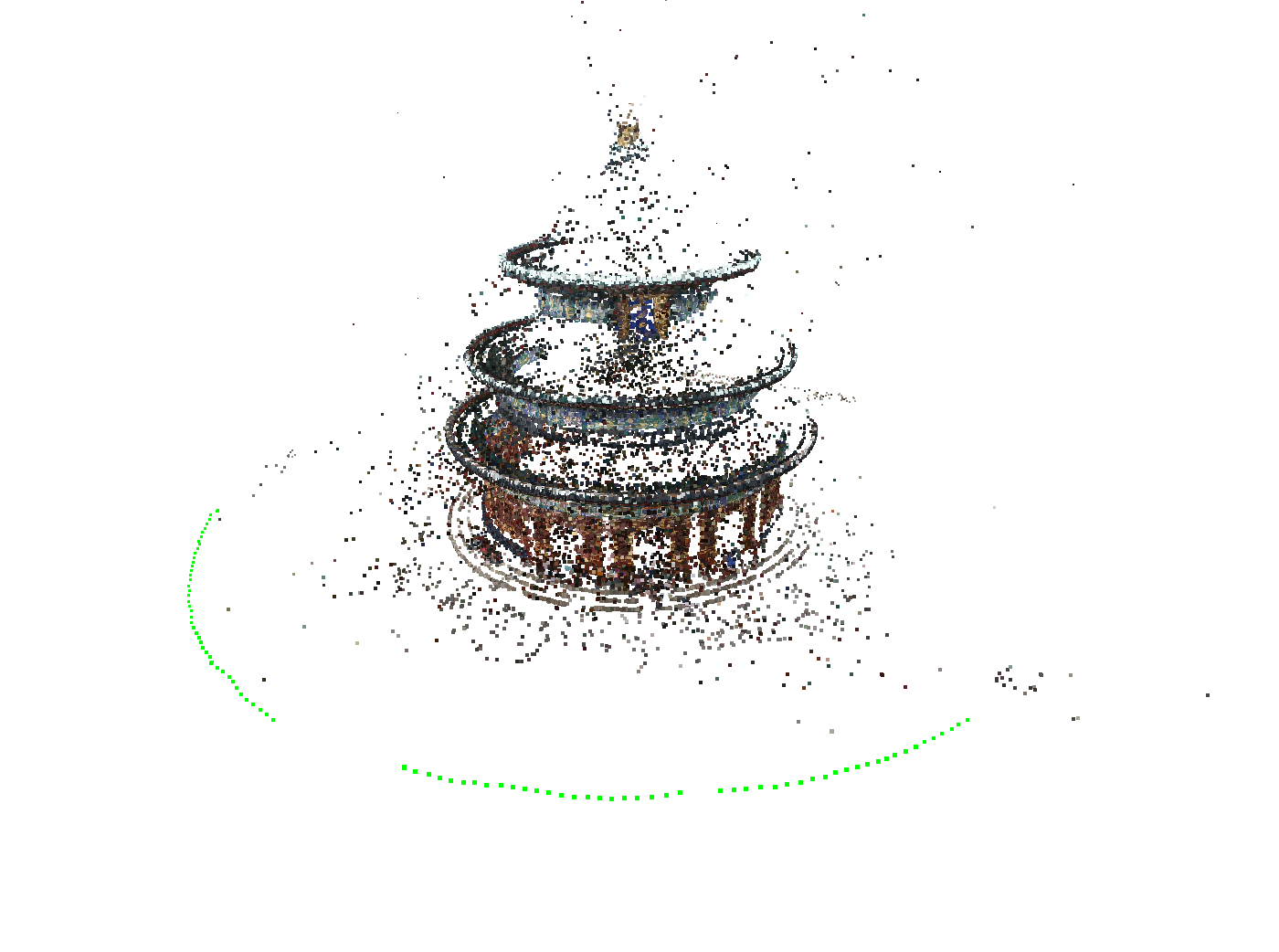}
        \caption{$t = 95$}
        \label{fig:temple60-95}
    \end{subfigure}
    \caption{
    Individual snapshots showing the viewgraph \emph{(top)} and the sparse structure \emph{(bottom)} of a reconstruction of the temple scene using a very unfortunate image ordering document the algorithm's recovery ability.
    Two separate parts of the temple are initially wrongly reconstructed in a single model (a-b) and are later successfully separated (c) and the real structure is recovered (d).
    \vspace{-1em}
    }
    \label{fig:experiment60}
\end{figure*}

Within the experiments we run the pipelines in different modes:
\begin{description}
  \item [batch mode]
  All images are added to the pipeline at once and the model is reconstructed without intermediate results.
  This is the classical \gls{sfm} operation mode.
  \item [progressive mode]
  Images are fed to the pipeline in a certain order and intermediate reconstructions are enforced. In VisualSFM this is realized by the ``Add Image'' option and in Theia we dictated the order in which views are added to an ongoing reconstruction by a slight code modification.
\end{description}

\begin{figure}[t]
  \centering
  \includegraphics[width=0.85\linewidth]{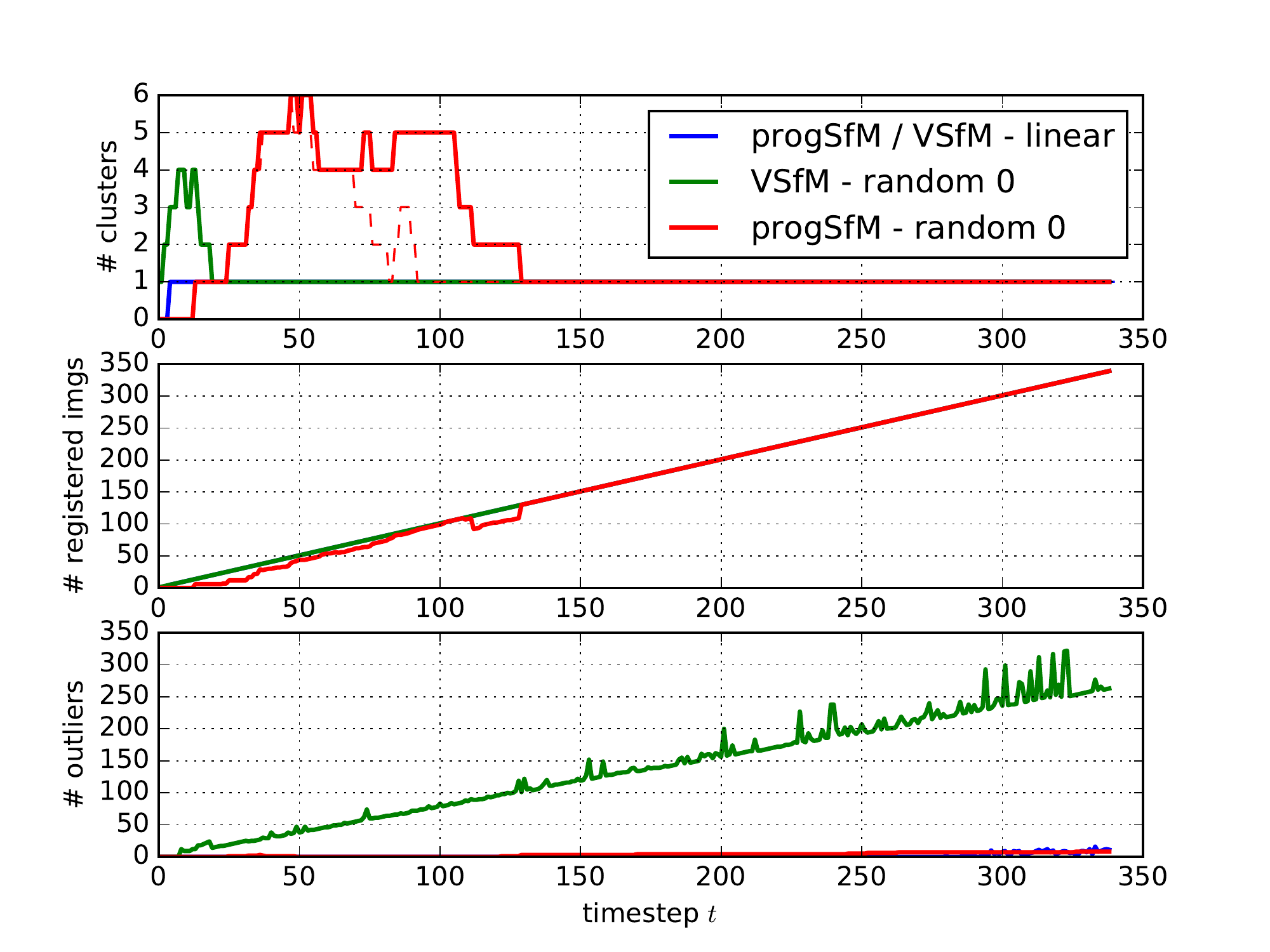}
  \caption{
    Behavior of the incremental and progressive \gls{sfm} pipelines on the TempleOfHeaven dataset.
    In the well behaving ``linear'' case \emph{(blue)}, both pipelines reconstruct the temple.
    If the input order is shuffled the incremental pipeline gets stuck in a local minimum \emph{(green)} whereas the proposed pipeline recovers and reconstructs the whole scene \emph{(red)}.
    \vspace{-1em}
  }
  \label{fig:temple-progressive}
\end{figure}

\subsection{Randomized Image Order}
One of the key contributions of the proposed pipeline is the ability of recovering from wrong connections
between image pairs made in previous reconstruction cycles.
Wrong connections often occure in symmetric environments which is why we chose the publicly
available TempleOfHeaven dataset~\cite{Jiang2012} with a rotationaly symmetric structure for the first experiment.
The dataset consists of 341 images taken in a regular spacing and perfectly demonstrates problems arising from unpleasant orderings.
As a reference model, we reconstructed the scene using VisualSFM and restricted the matches to sequential matching in the original order.
Figure~\ref{fig:temple-progressive} shows the development of the model as a function of the timestep $t$.
We report the number of clusters before and after global cluster position estimation as well as the number of outlier cameras.
A camera is considered to be an outlier if the position error is larger than half the minimum distance between two cameras in the reference model.

As expected both methods perform well in the ``linear'' case, where the images are fed to the pipeline in the original ordering.
A single cluster is reconstructed and maximum of 8 cameras are classified as outliers.
In a second experiment we randomly shuffled the order of images.
After an initial set of 20 images, the baseline merges all clusters to a single one and as a result the number of outlier cameras increases linearly.
The final 3D model is highly corrupted and only covers about a third of the temple (Figure~\ref{fig:temple100-vsfm}).

In contrast, the proposed pipeline creates up to 6 individual but locally consistent clusters.
After 84 out of the 341 images the clusters are correctly localized into a single effective cluster by the cluster positioning (dashed line).
Figure~\ref{fig:temple100} shows the resulting reconstruction after 100 images as well as the corresponding clustered viewgraph.

\begin{figure}[t]
\centering
\includegraphics[trim={2.5cm 1cm 1cm 2cm},clip,width=0.35\linewidth]{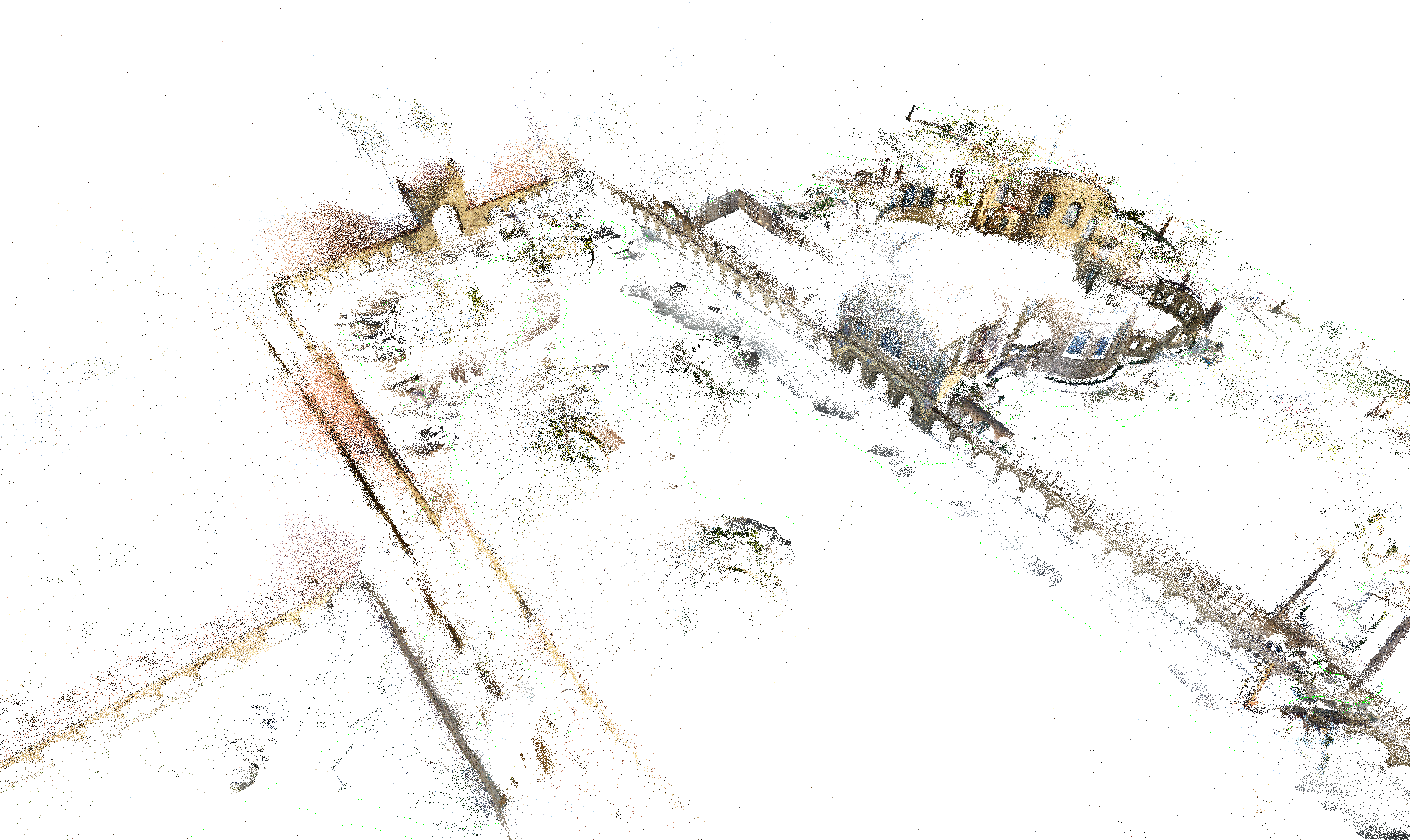}
\includegraphics[trim={2.5cm 1cm 1cm 2cm},clip,width=0.35\linewidth]{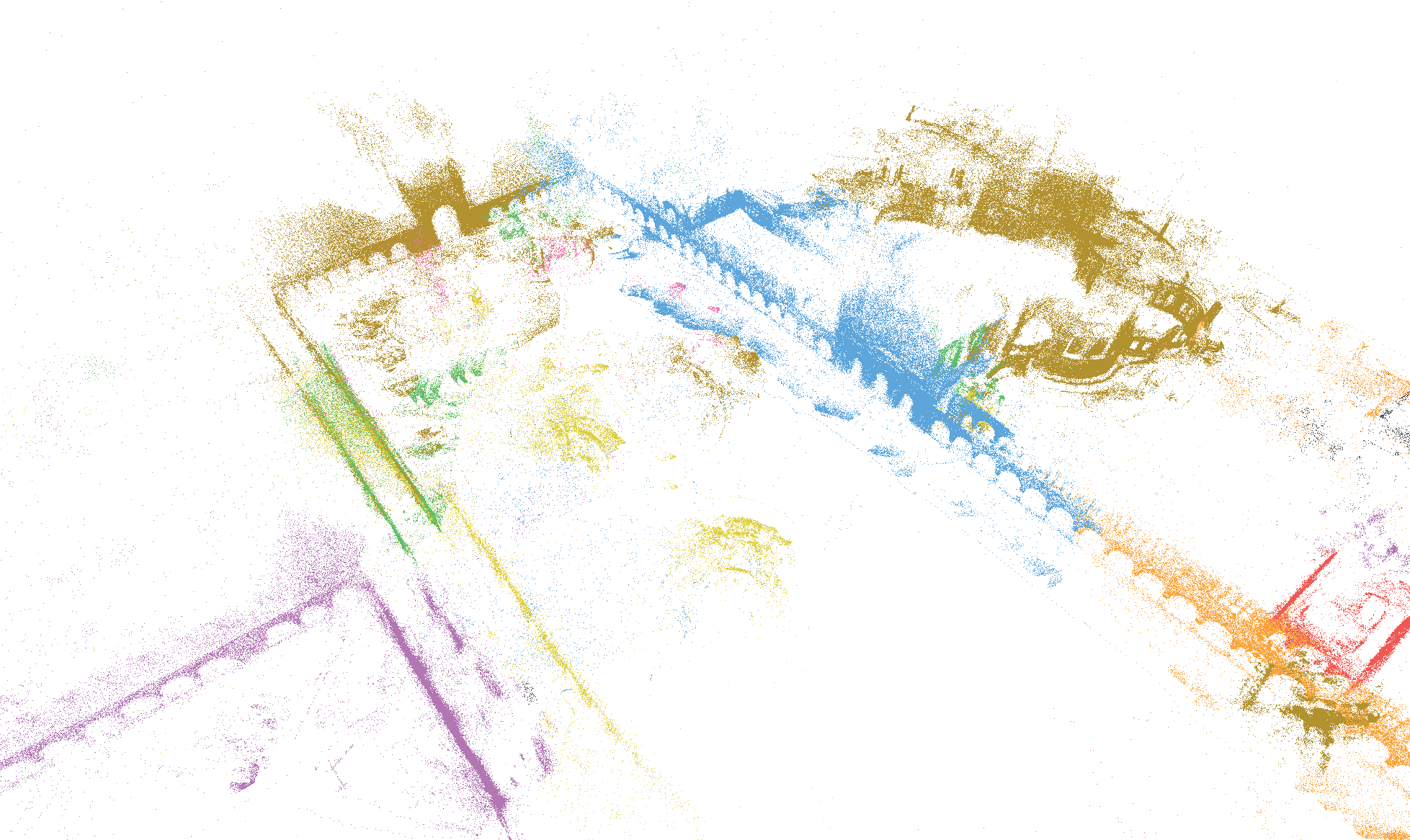}
\includegraphics[trim={4cm 0cm 3cm 3cm},clip,width=0.28\linewidth]{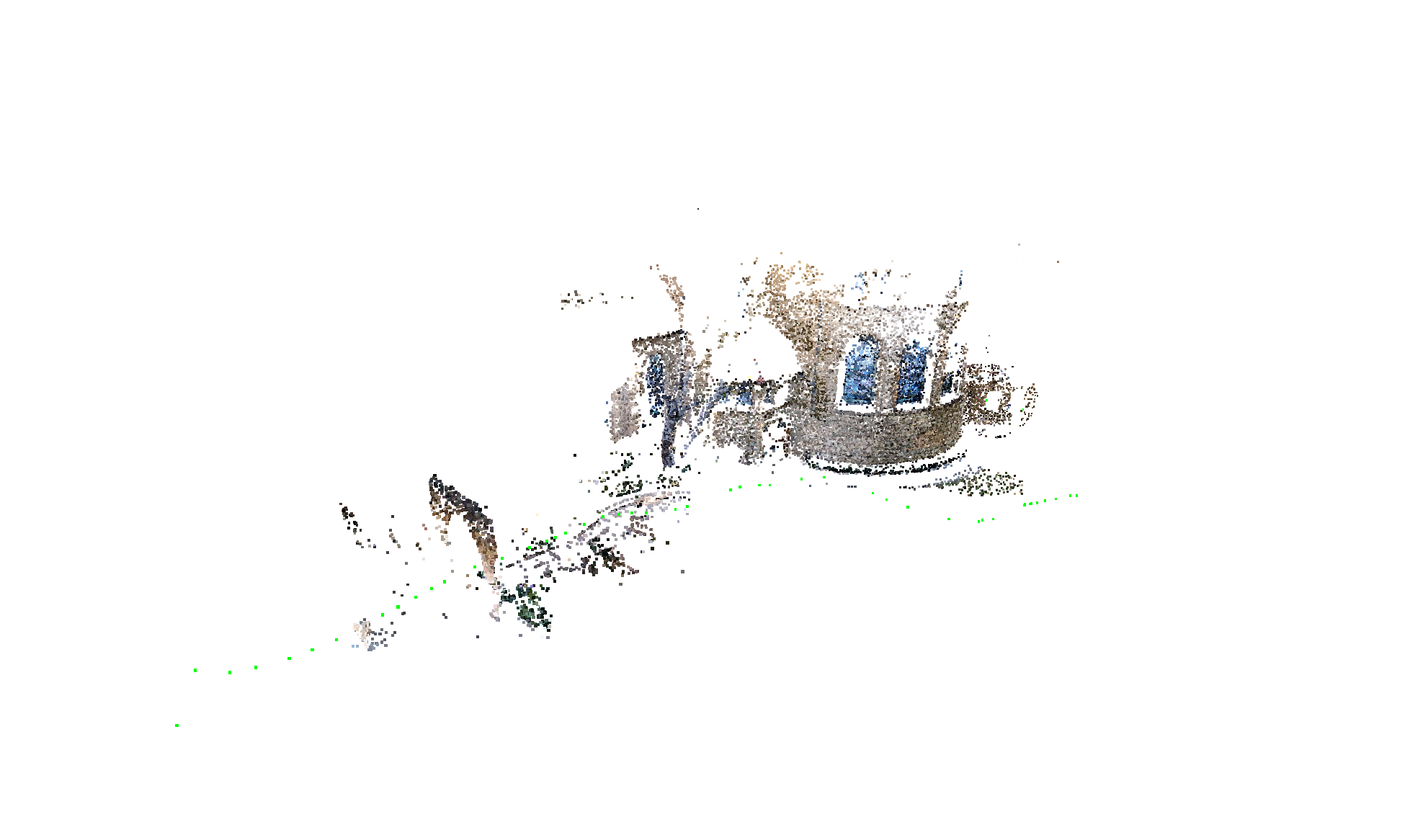}

\caption{Comparison of the proposed progressive pipeline \emph{(left, center)} to the incremental \gls{sfm} pipeline \emph{(right)} on the Stanford dataset. }
\label{fig:progrec_large}
\end{figure}

\subsection{Very Unfortunate Image Order}
\begin{figure}
  \centering
  \includegraphics[trim={1cm 0cm 1cm 0cm},clip,width=0.85\linewidth]{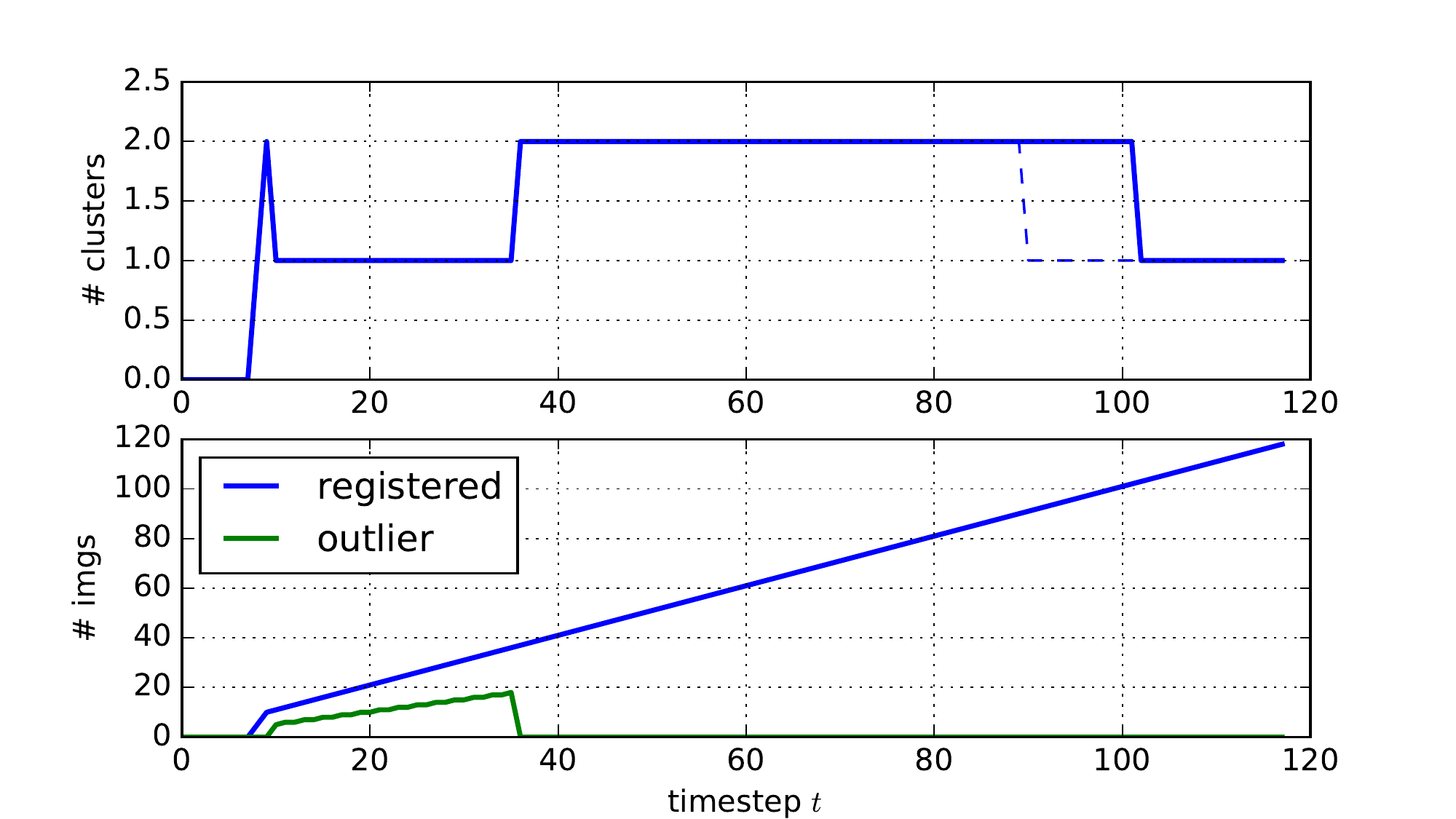}
  \caption{
    Evolution of the clusters and images in a very unfortunately ordered image sequence.
    Despite the heavy rotational symmetry of the temple dataset, the pipeline is able to recover from a wrong configuration (timestep 36).
    \vspace{-1em}
  }
  \label{fig:temple-plot-60}
\end{figure}

In order to demonstrate the recovery capabilities of our proposed pipeline we conducted an additional experiment on the TempleOfHeaven dataset.
The temple has a strong rotational symmetry which repeats with 60 degrees.
One of the worst conditions for an algorithm would be if the image ordering had exactly this 60 degree periodicity.
To test the algorithm, we created an artificial image sequence by feeding the images alternatively in the following order: $(I_0, I_{60}, I_{1}, I_{61},... )$.
Figure~\ref{fig:temple-plot-60} shows the number of clusters, images, and outliers in every timestep.
Due to the high amount of matches between the symmetric part and the lack of images in between, the clustering algorithm sees enough evidence for putting all images into a single cluster (Figure~\ref{fig:temple60-10}) and roughly every second image is an outlier.
After the addition of the 36th image, the connectivity has sufficiently changed so that the two clusters are recognized and the 3D~models are separated.

Due to the highly interleaved configuration the structure of the individual models cannot be recycled in this case and both clusters are reconstructed from scratch.
At this stage there are no inter-cluster constraints available yet, which is why the global cluster positioning does not succeed and the models are displayed as independent sub-models (Figure~\ref{fig:temple60-36}).
On the 95th image enough intra-cluster evidence is available s.t.\ the models can be placed into a common coordinate system (Figure~\ref{fig:temple60-95}) and with the 101st image a single cluster is formed (this time the structure of the individual sub-models can be recovered and a simple merging is needed).
The experiment shows that the proposed pipeline can recover from a wrong local minimum of the reconstruction.

\subsection{Realworld Progressive Reconstruction}
While the experiments so far demonstrated the capabilities of the progressive pipeline for unfortunate
image orderings, its behavior in real world applications remains to be shown.
Therefore we collected a series of 4516 images with three different mobile devices (\emph{HTC Nexus 9}, \emph{LG Nexus 5x}, and \emph{Google Pixel}) consisting of 29 sequences on the main quad of the Stanford University.
The sequences were then progressively reconstructed in an interleaved order (simulating multiple users
collaboratively acquiring the pictures) as well as in the batch-processing mode.
An additional similar experiment was conducted using the publicly available Quad dataset~\cite{disco2013pami} consisting of 6514 images, mainly taken by \emph{iPhone 3G}.
Images were shuffled randomly and pushed to the reconstruction algorithms.

Table \ref{tbl:comparison} shows the proposed pipeline compared to the incremental Theia~\cite{theia-manual} pipeline in batch and progressive modes.
In addition we compare against the HSfM~\cite{Cui2017} pipeline purely in batch-processing mode.
The proposed algorithm as the only pipeline can deal with the progressive scheme while the other get stuck in a local minimum and only reconstruct a very small subpart of the scene\footnote{Experiments were repeated using multiple random orderings.}.
In the case of the progressive pipeline, we also report the number of failure recoveries during the whole reconstruction (the number of resets with a local cluster due to major topological changes).
Figure~\ref{fig:progrec_large} illustrates the final result of the proposed pipeline versus the local minimum of the incremental pipeline.
The compute times of the proposed pipeline are comparable to existing pipelines despite the continuous flow of intermediate results.
This is due to the fact that our pipeline practically never operates on the whole image sets, but only on the local clusters.
This allows significantly faster execution times of the bundle adjustment.

\begin{figure}[t]
  \centering
  \adjincludegraphics[max width=0.30\linewidth, trim=200px 150px 50px 120px, clip=true]{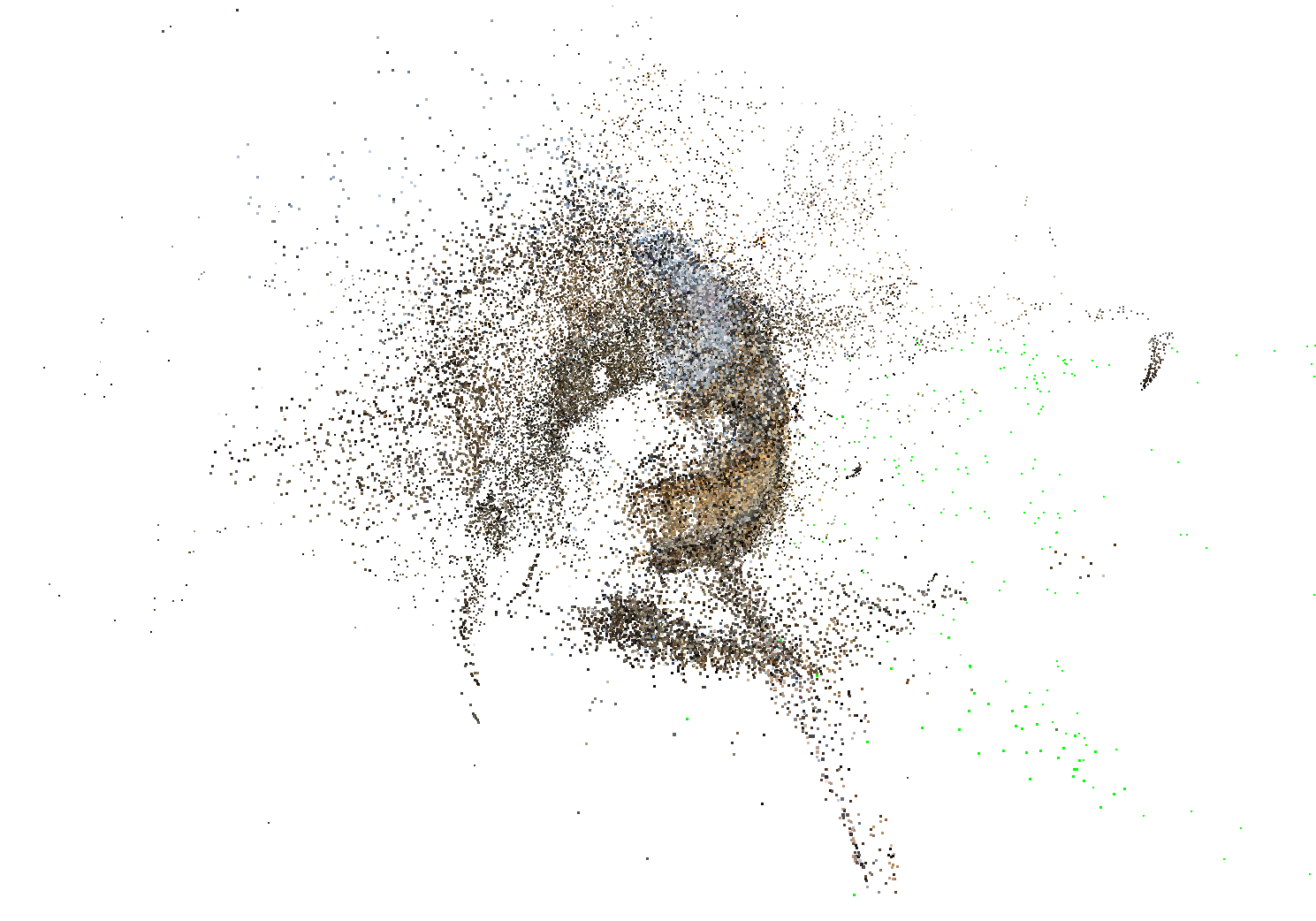}
  \adjincludegraphics[max width=0.30\linewidth, trim=200px 200px 200px 200px, clip=true]{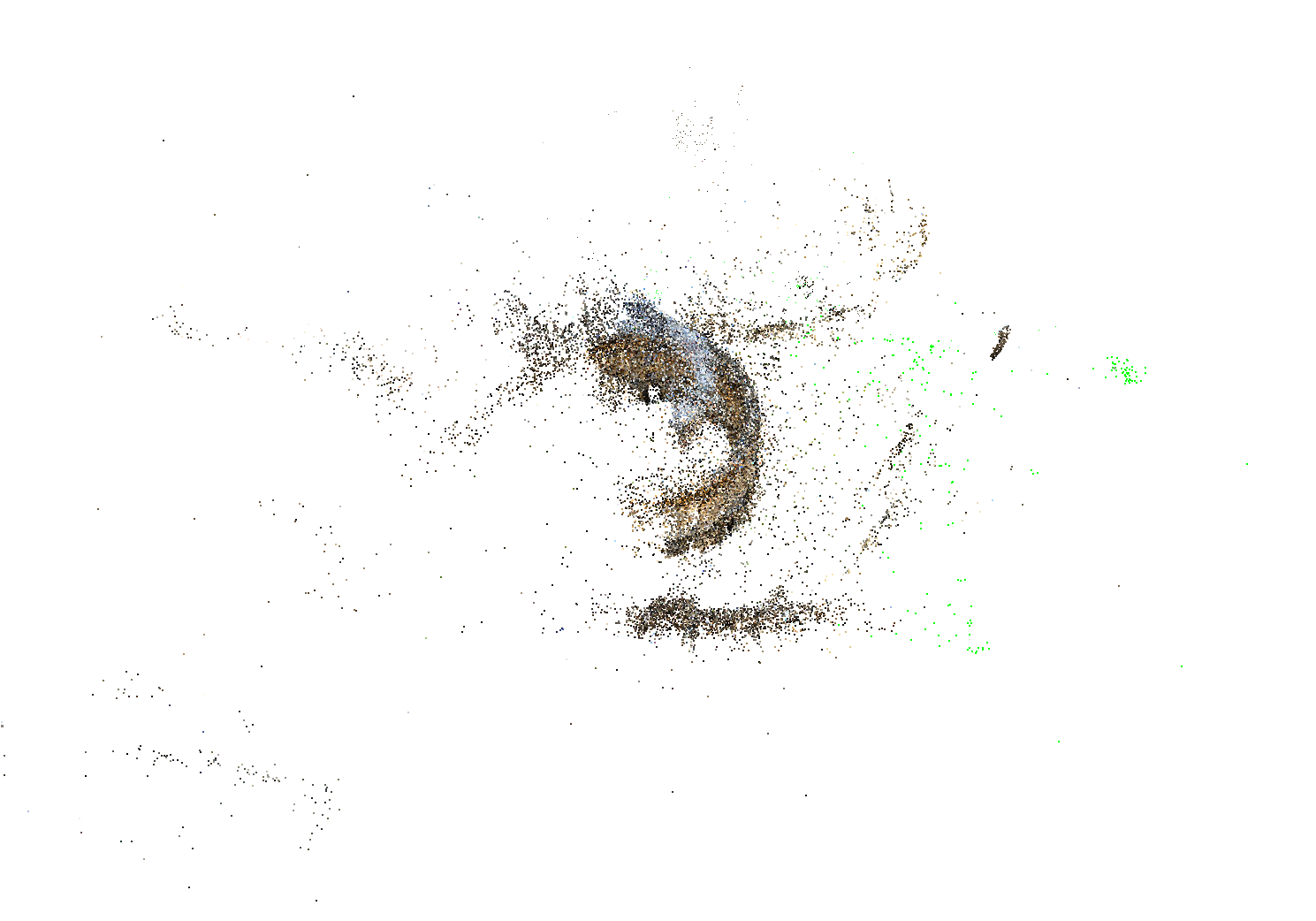}
  \adjincludegraphics[max width=0.35\linewidth, trim=200px 250px 180px 200px, clip=true]{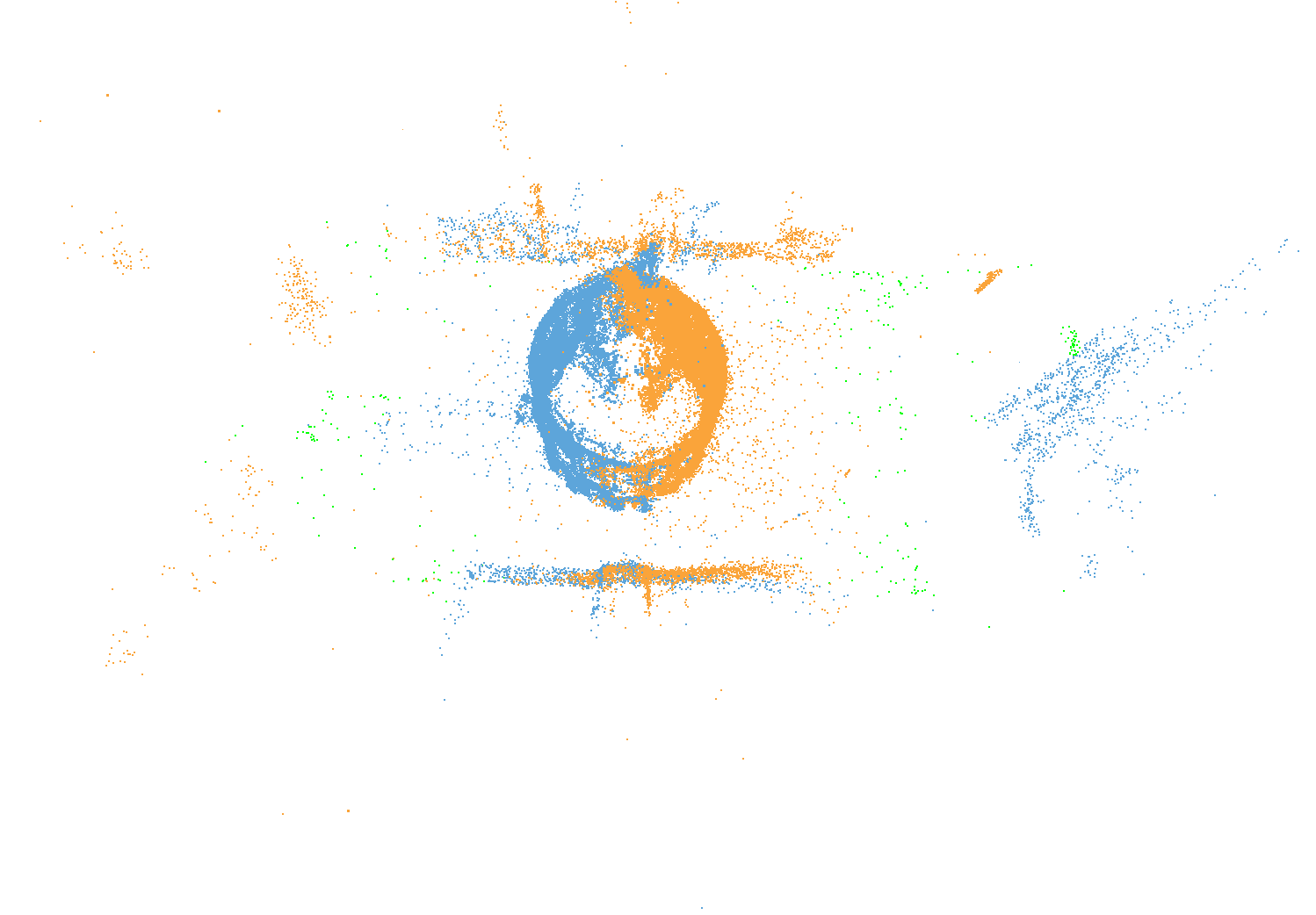}
  \caption{Result on the Radcliffe dataset with HSfM (\emph{left}), Theia (incr.)(\emph{right}) and the proposed method (\emph{right}). }
  \label{fig:duplicate}
\end{figure}

We furthermore run our method on several of the datasets presented by \cite{Heinly2014}.
We used random ordering of the crowd sourced data and pushed them to the progressive pipeline as done in the Quad~\cite{disco2013pami} experiment.
Figure~\ref{fig:duplicate} shows a sample result on the Radcliffe scene.
While both HSfM and the incremental baseline method got confused by the duplicate structure, our pipeline successfully separated the front and rear views of Radcliffe.
The two clusters were merged successfully into the correct configuration by the global cluster pose optimization.
In contrary to~\cite{Heinly2014}, this is not done in a post processing step but erroneous connections are detected on-the-fly during the reconstruction.
Table~\ref{tbl:comparison} shows the results on the resulting numbers on the dataset equally to the experiment before.
Our method was able to separate duplicate structure in all datasets.
As the reconstruction backbone of the presented pipeline bases on the well known incremental~\cite{theia-manual} and global~\cite{Wilson2014} pipelines, the accuracy of the resulting structure is equals the accuracy of these pipelines.

\begin{table}[t]
\centering
\begin{adjustbox}{width=0.95\linewidth}
\begin{tabular}{llllrrrrrrr}
\toprule
& & & & \multirow{3}{1.5cm}{Stanford} & \multirow{3}{1.5cm}{Quad~\cite{disco2013pami}} & \multirow{3}{1.5cm}{BigBen} & \multirow{3}{1.5cm}{Radcliffe Camera} & \multirow{3}{1.5cm}{Alexander Nevsky Cathedral} & \multirow{3}{1.5cm}{Branden- burg Gate} & \multirow{3}{1.5cm}{Arc De Triomphe} \\
\\
\\ \addlinespace[-0.3ex]
& & total imgs & [\#] & 4516 & 6514 & 402 & 282 & 448 & 175 & 434 \\ \addlinespace[-0.3ex]
\midrule
\multirow{4}{*}{ \rotatebox[origin=c]{90}{\parbox[c]{1cm}{\centering batch}} } & HSfM~\cite{Cui2017} & imgs & [\#] & 2,140 & 4,832 & 375 & 272 & 430 & 173 & 441 \\ \addlinespace[-0.3ex]
& & time & [s] & 11,490 & 23,570 & 1,932 & 1,654 & 1,900 & 389 & 2,131 \\ \addlinespace[-0.3ex]
\cmidrule(lr){2-11}
& Theia (incr.)~\cite{theia-manual} & imgs & [\#] & 3,298 & 5,462 & 394 & 278 & 443 & 173 & 410 \\ \addlinespace[-0.3ex]
& & time & [s] & 34,749 &  158,853 & 2,385 & 1,307 & 3,687 & 1,018 & 3,348\\ \addlinespace[-0.3ex]
\midrule
\multirow{7}{*}{ \rotatebox[origin=c]{90}{\parbox[c]{1cm}{\centering progressive}} } & Theia (incr.)~\cite{theia-manual} & imgs & [\#]& 51 & 527 & 394 & 280 & 418 & 173 & 416 \\ \addlinespace[-0.3ex]
& & time & [s] & 7,605 & 215,064 & 2,026 & 1,473 & 2,268 & 1,141 & 3,516 \\ \addlinespace[-0.3ex]
\cmidrule(lr){2-11}
& proposed & imgs & [\#] & 3,165 & 3,894 & 285 & 279 & 427 & 173 & 405 \\ \addlinespace[-0.3ex]
& & time & [s] & 13,276 & 25,713 & 552 & 1,121 & 2,163 & 627 & 377 \\ \addlinespace[-0.3ex]
& & clusters & [\#] & 76 & 92 & 5 & 2 & 4 & 2 & 5\\ \addlinespace[-0.3ex]
& & recoveries & [\#] & 14 & 29 & 1 & 0 & 5 & 1 & 1 \\ \addlinespace[-0.3ex]
\bottomrule
\end{tabular}
\end{adjustbox}
\vspace{0.3em}
\caption{
Evaluation of our pipeline versus the incremental Theia pipeline and HSfM. Only the proposed pipeline  can successfully handle the progressive scenario.
\vspace{-1em}
}
\label{tbl:comparison}
\end{table}

\vspace{-0.3em}
\section{Conclusions}
We proposed a novel progressive \gls{sfm} pipeline which addresses a multiuser-centric scenario, where a 3D~model is simultaneously reconstructed from multiple image streams handled by a cloud-based reconstruction service.
In contrary to existing work, our pipeline does not depend on the image order and does not require any a-priori global knowledge about image connectivity.
The progressive pipeline avoids taking any binding decisions and is able to recover for erroneous configurations. 
A global viewgraph is incrementally built and maintained.
The graph is clustered based on the local connectivity of cameras and individual clusters are reconstructed using either an incremental or a global reconstruction pipeline.
In the last step, individual models are merged using a lightweight posegraph optimization just on the cluster centers.
We demonstrated the effectiveness and efficiency of our pipeline on multiple dataset and compared it to existing solutions.

{\small
\bibliographystyle{ieee}
\bibliography{egbib}
}

\end{document}